\documentclass[letterpaper]{article} % DO NOT CHANGE THIS
\usepackage[submission]{aaai23}  % DO NOT CHANGE THIS
\usepackage{times}  % DO NOT CHANGE THIS
\usepackage{helvet}  % DO NOT CHANGE THIS
\usepackage{courier}  % DO NOT CHANGE THIS
\usepackage[hyphens]{url}  % DO NOT CHANGE THIS
\usepackage{graphicx} % DO NOT CHANGE THIS
\usepackage{natbib}  % DO NOT CHANGE THIS AND DO NOT ADD ANY OPTIONS TO IT
\usepackage{caption} % DO NOT CHANGE THIS AND DO NOT ADD ANY OPTIONS TO IT
\usepackage{algorithm}
\usepackage{algorithmic}
\usepackage{amsmath}
\usepackage{amssymb}
\floatname{algorithm}{Algorithm}

\usepackage{newfloat}
\usepackage{listings}
\DeclareCaptionStyle{ruled}{labelfont=normalfont,labelsep=colon,strut=off} % DO NOT CHANGE THIS
\floatstyle{ruled}
\newfloat{listing}{tb}{lst}{}
\floatname{listing}{Listing}
\usepackage{booktabs}
\usepackage{multirow}
\usepackage{xcolor}

\usepackage{lipsum}
\usepackage{tabularx}
\usepackage{booktabs}
\usepackage{multirow}
\usepackage{arydshln}
\usepackage{pgfplots}

\title{Multi-modal Latent Space Learning 
for Chain-of-Thought Reasoning in Language Models}
\author{Liqi He$^{1,2,\dag}$, Zuchao Li$^{1,2,\dag}$\thanks{$\ $  Corresponding author. $^\dag$ Equal contribution. This work was supported by the National Natural Science Foundation of China (No. 62306216), the Natural Science Foundation of Hubei Province of China (No. 2023AFB816), the Fundamental Research Funds for the Central Universities (No. 2042023kf0133), National Natural Science Foundation of China [No. 72074171] [No. 72374161]. }, Xiantao Cai$^{1}$, and Ping Wang$^{3,4}$\\
$^{1}$National Engineering Research Center for Multimedia Software, \\
School of Computer Science, Wuhan University, Wuhan, 430072, China \\
$^{2}$Hubei Luojia Laboratory, Wuhan 430072, China \\
$^{3}$Center for the Studies of Information Resources, Wuhan University, Wuhan 430072, China\\
$^{4}$School of Information Management, Wuhan University, Wuhan 430072, China
{\tt \{heliqi,zcli-charlie,caixiantao,wangping\}@whu.edu.cn}\\
}

\usepackage{bibentry}
\begin{document}

\maketitle

\begin{abstract}
Chain-of-thought (CoT) reasoning has exhibited impressive performance in language models for solving complex tasks and answering questions. However, many real-world questions require multi-modal information, such as text and images. Previous research on multi-modal CoT has primarily focused on extracting fixed image features from off-the-shelf vision models and then fusing them with text using attention mechanisms. This approach has limitations because these vision models were not designed for complex reasoning tasks and do not align well with language thoughts. To overcome this limitation, we introduce a novel approach for multi-modal CoT reasoning that utilizes latent space learning via diffusion processes to generate effective image features that align with language thoughts. Our method fuses image features and text representations at a deep level and improves the complex reasoning ability of multi-modal CoT. We demonstrate the efficacy of our proposed method on multi-modal ScienceQA and machine translation benchmarks, achieving state-of-the-art performance on ScienceQA. Overall, our approach offers a more robust and effective solution for multi-modal reasoning in language models, enhancing their ability to tackle complex real-world problems.
\end{abstract}

\section{Introduction}

\begin{figure}[t]
    \centering
    \includegraphics[width=.45\textwidth]{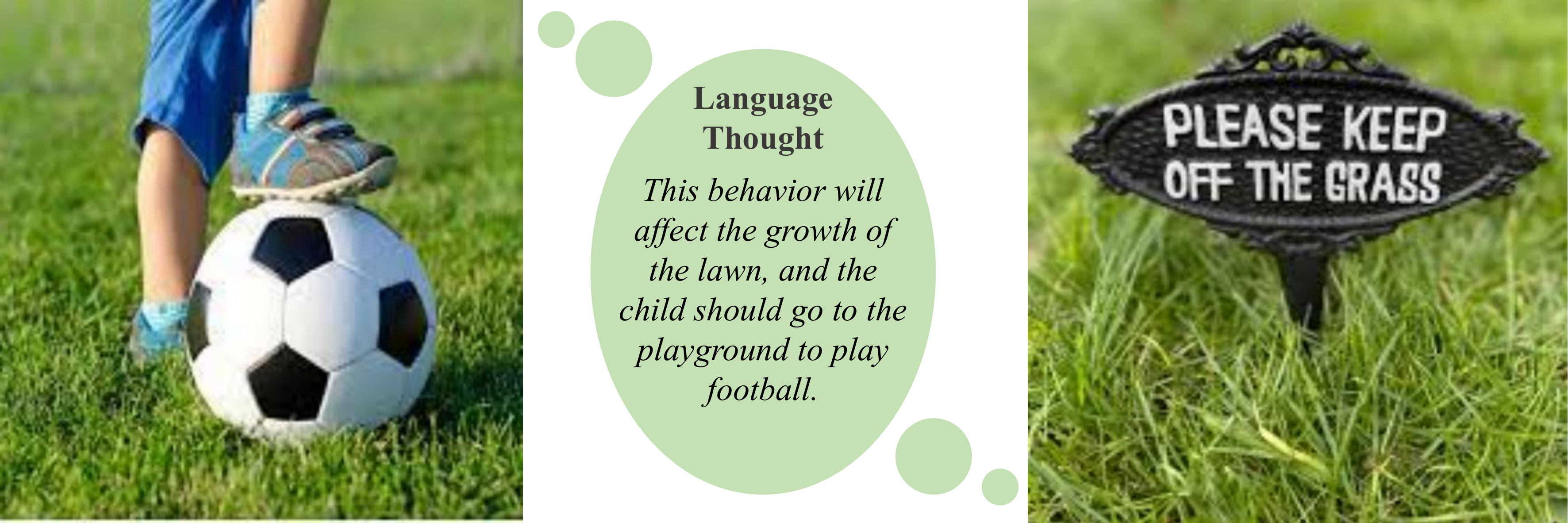}  
    \caption{Language Thought
    % The figure illustrates a scenario where a child is playing football on a lawn in a park, while a sign saying "Please keep off the grass" is also present. In this situation, our (language) thoughts may turn towards the potential negative impact of the child's behavior on the grass.
    }
    \label{fig:example}
    \vspace{-6mm}
\end{figure}

In our daily lives, we are constantly bombarded with information from various sources, such as text, images, and more. To make sense of this complex world, we need to be able to acquire and integrate multi-modal information effectively. For example, as shown in Figure \ref{fig:example}, when we see the slogan "Please keep off the grass"
% (\textcolor{red}{language modality})
(language modality)
on the lawn of the park and a child is playing football on the same lawn
% (\textcolor{red}{visual modality})
(visual modality)
, we think of the negative effects of trampling the grass on the park's ecological environment and prepare to told the child to go to the football field
% (\textcolor{red}{language thought})
(language thought)
. These ideas come from our deep understanding and reasoning of linguistic and visual information, which can be called language thought.

In recent years, chain-of-thought (CoT), which involves a series of intermediate reasoning steps (also known as rationale), has significantly enhanced the complex reasoning ability of large language models by providing them with access to a portion of language thought ~\cite{DBLP:journals/corr/abs-2201-11903}. 
% For example, CoT of large language models dramatically improve the performance of large language models on arithmetic problems and symbolic reasoning. Existing CoT prompting can be categorized into two major paradigms: Zero-shot-CoT (Figure \ref{fig:introduction} (a)) and Few-shot-CoT (Figure \ref{fig:introduction} (b)). Zero-shot-CoT~\cite{DBLP:journals/corr/abs-2205-11916} leverages a single prompt like “Let’s think step by step” to generate reasoning chains. Few-shot-CoT~\cite{DBLP:journals/corr/abs-2201-11903} uses reasoning demonstrations one by one. For example, prompting a PaLM 540B with eight chain-of-thought exemplars achieved state-of-the-art accuracy on the GSM8K benchmark of math word problems, surpassing even finetuned GPT-3 with a verifier~\cite{DBLP:journals/corr/abs-2201-11903}. 

\begin{figure}
    \centering
    \includegraphics[width=.45\textwidth]{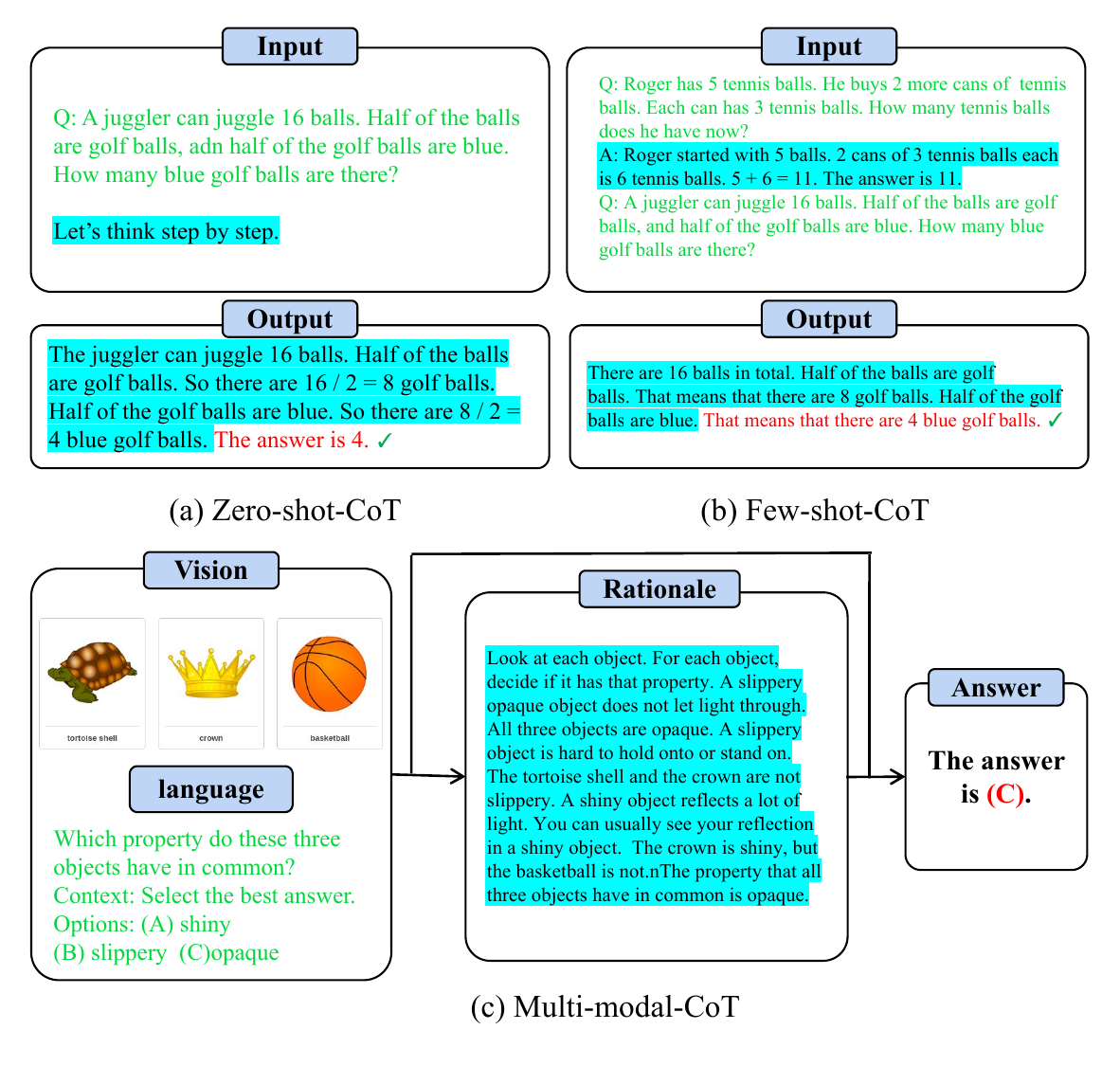}
    \vspace{-3mm}
    \caption{(a) Zero-shot-CoT~\cite{DBLP:journals/corr/abs-2205-11916}  (b) Few-shot-CoT~\cite{DBLP:journals/corr/abs-2201-11903} 
    (c) Multi-modal-CoT~\cite{DBLP:journals/corr/abs-2302-00923}}
    \label{fig:introduction}
    \vspace{-6mm}
\end{figure}

Training smaller language-only models with less than 100 billion parameters for CoT reasoning remains a significant challenge due to hallucination and tend to produce illogical rationales. 
% This is particularly challenging because 
% Real-world problems often require the integration of information from multiple sources, such as text and images. 
To address these problems more effectively, it is crucial to enable large language models to develop a deeper understanding of multi-modal information and generate more effective language thought.
One solution that has been proposed to help integrate information across visual and linguistic modalities is Multi-Modal CoT (MM-CoT)~\cite{DBLP:journals/corr/abs-2302-00923}.
% , which takes input from both images and text. 
% However, leveraging multi-modal information for language models can also be difficult. 
% To overcome this challenge, 
MM-CoT extracts fixed image features and text representations and fused them to obtain multi-modal features. 
% These models usually 
MM-CoT adopts a two-stage framework that includes rationale generation and answer inference, as shown in Figure \ref{fig:introduction} (c). This approach has been shown to outperform generating rationale and answer together on question answering tasks~\cite{DBLP:journals/corr/abs-2302-00923}.
% Experiments have shown that vision features play a crucial role in helping language models generate high-quality rationales, which in turn leads to more accurate answers. In fact, MM-CoT outperforms ChatGPT significantly and even surpasses human performance on the ScienceQA benchmark~\cite{DBLP:journals/corr/abs-2302-00923}.

However, existing multi-modal CoT models rely on fixed image features extracted by pre-trained vision extraction models such as DETR~\cite{DBLP:conf/eccv/CarionMSUKZ20} or CLIP~\cite{DBLP:conf/icml/RadfordKHRGASAM21}. However, fixed image features do not align well with flexible text queries. And vision models that extract these features are not optimized for producing useful visual information that would lead to effective rationales generated by language models. 
For example, 
% while DETR detects objects, its extracted features may only pay attention to the main objects in an image. Additionally, 
while CLIP is trained on (image, text) pairs, it only extracts shallow image information. Shallow vision features may not help language models infer correct answers because they are not closed to the reasoning. For example, as shown in Figure \ref{fig:example}, we can not synthetize language thought if we look at the lawn in the picture and the text on the banner separately. We hypothesize that in both the stage of rationale generation and the stage of answer inference for complex problem solving, there is a need for deep understanding of visual features that capture different information in images.
Therefore, effectively utilizing different modalities remains a key challenge. In this work, we propose an approach to enhance the complex reasoning ability of large language models by improving their ability to synthesize and employ language thoughts. Our approach leverages both language and vision information to achieve this goal. We propose to obtain a multi-modal latent space that deeply fuses visual features and text representations via a diffusion process. This allows our method to develop deep-level understanding, alignment and reasoning of both visual and linguistic modalities, resulting in more effective language thought generation.

% \textcolor{red}{
% (Delete the introduction to diffusion models in the introduction)}

Drawing inspiration from diffusion models(more information on diffusion models in the Appendix, we employ the diffusion process to learn a text-image aligned latent space for language thought reasoning. 
The diffusion process entails the sequential application of multiple transformations to the latent space of image representation, where the level of noise is gradually augmented with each iteration. As a result, a series of increasingly blurred representations of the original image input is generated, ultimately leading to random noise that follows a Gaussian distribution. During each stage of the noise prediction, the model acquires a novel representation of the joint text-image distribution that captures more intricate dependencies and higher-level semantics. By repeating this procedure across several iterations, the model can acquire a deep and well-aligned latent space that encodes abundant information about both modalities.
This approach is particularly useful for CoT reasoning tasks, where the goal is to reason about a long sequence of inputs and their corresponding image. By learning a deep latent space that captures high-level dependencies between text and images, it is well-suited for complex reasoning tasks.

We conducted experiments on the ScienceQA benchmark, which contains questions that require reasoning based on provided text and images. The results show that our proposed latent space learning is effective in generating useful chain of thought (CoT) and inferring correct answers. We achieved new state-of-the-art results on the ScienceQA benchmark with about only 1 billion parameters, outperforming the current SOTA baseline by 6.06\% (base), 1.67\% (large) respectively, and the strong ChatGPT system by 18.18\% with less than 1/100th of the parameters, demonstrating the effectiveness of our approach. Our method also demonstrates strong ability in generating effective CoT, as evidenced by the ROUGE-L score of the rationales outperforming the baseline by 1.21.
In addition to ScienceQA, we evaluated the effects of diffusion process for multi-modal latent space learning in multi-modal machine translation, where it also brought significant improvements. These results suggest that our proposed method is a general enhancement and can benefit the multi-modal information processing community.

\begin{figure*}[t]
  \centering
  \includegraphics[width=0.9\textwidth]{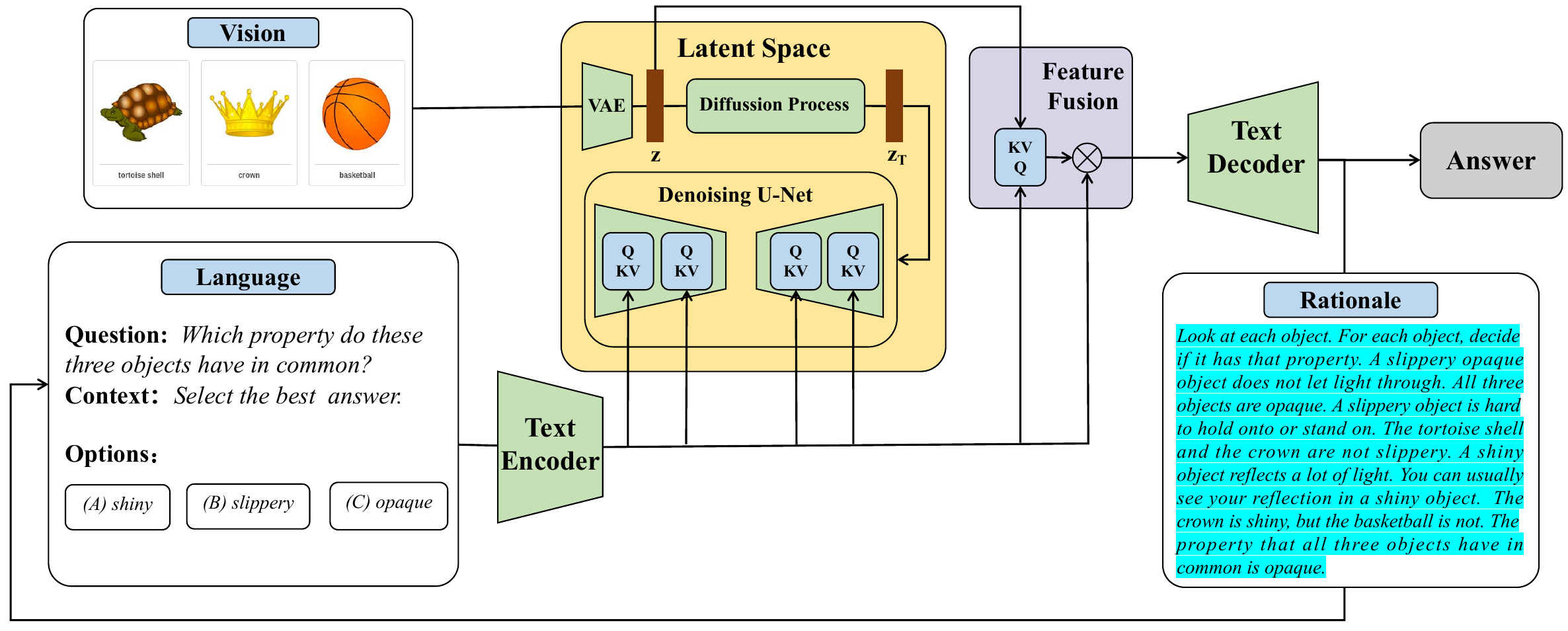}
  \vspace{-3mm}
  \caption{Overview of our multi-modal latent space learning via diffusion process for chain-of-thought reasoning in language models. Our framework consists of two stages: (i) rationale generation and (ii) answer inference.}
  \label{fig:overview}
  \vspace{-4mm}
\end{figure*}

\section{Related Work}

\subsection{CoT Reasoning in LLMs}

CoT is a widely applicable method for enhancing complex reasoning in large language models (LLMs)~\cite{DBLP:journals/corr/abs-2201-11903}. CoT techniques assist LLMs in generating a series of logical reasoning steps, enabling them to think step by step about a question and arrive at the correct answer. CoT has significantly improved language models' performance in generating rationales and inferring accurate answers in numerous domains, including commonsense and arithmetic. In this section, we will discuss the progress made in eliciting CoT reasoning by prompting and fine-tuning language models.

For example, CoT of large language models dramatically improve the performance of large language models on arithmetic problems and symbolic reasoning. Existing CoT prompting can be categorized into two major paradigms: Zero-shot-CoT (Figure \ref{fig:introduction} (a)) and Few-shot-CoT (Figure \ref{fig:introduction} (b)). Zero-shot-CoT~\cite{DBLP:journals/corr/abs-2205-11916} leverages a single prompt like “Let’s think step by step” to generate reasoning chains. Few-shot-CoT~\cite{DBLP:journals/corr/abs-2201-11903} uses reasoning demonstrations one by one. For example, prompting a PaLM 540B with eight chain-of-thought exemplars achieved state-of-the-art accuracy on the GSM8K benchmark of math word problems, surpassing even finetuned GPT-3 with a verifier~\cite{DBLP:journals/corr/abs-2201-11903}.

\subsection{Multi-Modal CoT Reasoning in LLMs}

To address the issue of hallucinations, which can lead to incorrect answers, and handle real-world multi-modal tasks effectively, multi-modal information can guide models to generate logical rationales. Recent studies on multi-modal Chain-of-Thought (CoT) outperform the previous state-of-the-art large language model (ChatGPT, GPT-3.5) by 16 percentage points, achieving 91.68\% accuracy and even surpassing human performance on the ScienceQA benchmark ~\cite{DBLP:journals/corr/abs-2302-00923}. 
% This approach uses text representation encoded by a large language model and fixed image features encoded by off-the-shelf vision extraction models, fused via an attention mechanism. In its implementation, it adopts a two-stage framework consisting of rationale generation and answer inference. Experimental results demonstrate that this framework performs better on question answering tasks than generating rationales and answers together.

Leveraging vision information effectively and fusing visual features with text representation in multi-modal Chain-of-Thought (CoT) poses a significant challenge. Prior work~\cite{DBLP:journals/corr/abs-2209-09513} has attempted to use image captions and incorporate them after text input, but this approach results in substantial information loss of images. Other studies have proposed a method that encodes texts and images using a Transformer encoder and convolutional neural network, respectively ~\cite{DBLP:journals/corr/abs-2301-03344}. The two sequences of representations are then fused using an attention layer for cross-modal interaction. To extract image features, Zhang et al.~\cite{DBLP:journals/corr/abs-2302-00923} employed off-the-shelf vision extraction models such as DETR~\cite{DBLP:conf/eccv/CarionMSUKZ20} or CLIP~\cite{DBLP:conf/icml/RadfordKHRGASAM21} to obtain patch-level features and fused the information from the two modalities using an attention mechanism.
For more information on CoT related work, please refer to the Appendix. \ref{sec:CoT} \ref{sec:mmCoT}

% Despite recent advances, there are still challenges in using the visual modality more effectively. Firstly, most off-the-shelf vision extraction models are not specifically trained for complex CoT reasoning. Secondly, when a two-stage framework is used, fixed image features may not be suitable for different stages and flexible language queries. It is our belief that fixed and shallow vision features may not be the optimal choice to enhance language models' complex reasoning abilities and performance on question answering tasks. As shown in Figure \ref{fig:example}, the shallow vision features just include information about grass, football, slogan sign and so on. But we need language models to generate more complex reasoning such as "This child should play football on the playground". So our work aims at learning flexible and deep vision features used to improve complex reasoning ability.

\section{Method}

% In this section, we will first introduce the concrete methodology of our model. Then we will analyse multi-modal latent space learning in detail. Lastly, we will introduce how diffusion process provide guidance for the model to produce latent space which fuses multi-modal information in deep level. The overview of our model is shown in Figure \ref{fig:overview}.
In this section, we introduce our proposed Diffusion Process enhanced Multi-Modal CoT (DPMM-CoT) method. We follow the Multi-Modal CoT (MM-CoT) approach proposed by Zhang et al. \cite{DBLP:journals/corr/abs-2302-00923} as our baseline. The overview of our full model is illustrated in Figure \ref{fig:overview} and consists of four components: Text Encoder, Latent Space Learning Module, Feature Fusion Module, and Text Decoder. Training for DPMM-CoT follows the same two-stage framework as the baseline MM-CoT, including rational generation and answer inference, with latent space learning playing a role in both stages. The detailed process is shown in Algorithm \ref{alg:method} in Appendix.

% Baseline: mmCoT
\subsection{Multi-modal CoT}

% Our model aims at leveraging multi-modal information more effectively and lead the language model with 1B parameters to generate rationale and infer answers. We adopt two-stage framework including rationale generation and answer inference. The two stages share same model architecture. The following part will concretely show four main parts of the model architecture: text encoder, image encoder, attention mechanism, decoder.
Our full model aims to leverage multi-modal information more effectively, leading the language model to generate rationales and infer answers. The two-stage framework includes rationale generation and answer inference, which share the same model architecture and training process.

\subsubsection{Task Definition}

% First of all, we introduce the task definition briefly. Given the language input $X_{language}$ and the vision input $X_{vision}$, we firstly encode $X_{language}$ and $X_{vision}$. We get text representations $Z_{language}$ by text encoder of T5 model. To obtain image features $Z_{vision}$, we leverage the Variational AutoEncoder and the Unet neural network with the idea of stable diffusion.
In multi-modal reasoning, a language input including a question $X_Q$, its language context $X_L$, the options $X_O$, and its corresponding image $X_V$ are usually given as inputs to the model. The model is required to answer a question $X_Q$ according to the options $X_O$ to obtain the answer $Y_A$ . In other words, the model is trained to maximize the likelihood between the predicted answer $\hat{Y_A}$ and the true answer $Y_A$ distribution. MM-CoT breaks down this problem into two steps through the introduction of a rationale $Y_R$: rationale generation and answer inference. In the rationale generation step, the model is required to predict a rationale $\hat{Y_R}$ that can infer the answer, which maximizes the likelihood between predictions and the standard rationale $R$ distribution. Then, in the second stage, based on the rationale $\hat{Y_R}$, along with the language input including the question $X_Q$, its language context $X_L$, the options $X_O$, and the corresponding image $X_V$, the model predicts the final answer $Y_A$.

\subsubsection{Text Encoder}

% The text input is different between two stages. In the stage of rationale generation, the text input includes question text, context text and multiple options. In the stage of answer inference, the text input includes question text, context text, multiple options and rationale generated from the first stage. We adopt Transformer model initialized by UnifiedQA\cite{khashabi-etal-2020-unifiedqa}. We get $Z_{language}$ by LanguageEncoder(Formula \ref{equ:language_encoder}):
For the multi-modal CoT tasks, the text input differs between the two stages. In the stage of rationale generation, the text input includes language context $X_L$, a question $X_Q$, and multiple options $X_O$. In the stage of answer inference, the text input comprises language context $X_L$, a question $X_Q$, multiple options $X_O$, and rationale $\hat{Y_R}$ generated from the first stage. We adopt the Transformer model for text encoding, which is initialized by the pre-trained model UnifiedQA~\cite{khashabi-etal-2020-unifiedqa}. We obtain the text representation $Z_{L}$ as follows:
\begin{align}
    Z^R_{L} &= \textsc{Encoder}_{text}([X_L; X_Q; X_O]),\\
    Z^A_{L} &= \textsc{Encoder}_{text}([X_L; X_Q; X_O, \hat{Y_R}]),
    \label{equ:language_encoder}
\end{align}
where $Z_L = Z^R_{L}$ in the stage of rationale generation and $Z_L = Z^A_{L}$ in the stage of answer inference.

\subsubsection{Image Feature Extraction and Feature Fusion}

In multi-modal CoT, the image encoder plays a crucial role in the CoT process as it helps to provide additional context and information to the model. By incorporating visual features extracted from input images, the model gains a better understanding of the overall context. Specifically, the image feature $Z_{V}$ is first extracted by an image encoder:
\begin{equation}
    Z_{V} = \textsc{Encoder}_{img}(X_{V}).
    \label{equ:image_encoder}
\end{equation}
Based on the acquired image features, in order to integrate the image and text encoding features, a linear layer is first used to map the image features. This is primarily for two purposes: to unify the dimensions of image and text features, and to project the image features onto the same feature space that can be fused with text features.
\begin{equation}
    Z^T_{V} = W_h * Z_{V},
\end{equation}
where $W_h$ is the learnable weight matrix.

% \subsubsection{Attention Mechanism}

% To make text representation and image features relevant, we adopt a single-head attention mechanism. The query is $Z_{language}$. The key is $Z_{vison}$. The value is $Z_{vision}$. The calculation formula is as follows(Equation \ref{attention_mechanism}):
As the image features and text features have different temporal lengths, we use an attention mechanism to project the image features onto the length of the text features based on the correlation between the image and text features. Specifically, we use $Z_L$ as the attention query , with $Z^T_V$ as attention keys and values. The resultant projected image features are as follows:
\begin{equation}
    Z_{V}^{attn} = Softmax(\frac{\mathcal{Q}\mathcal{K}^{T}}{\sqrt{d_k}})\mathcal{V}
    \label{attention_mechanism}
\end{equation}
where $d_k$ is the dimension of $Z_L$, $\mathcal{Q}$ is the $Z_L$, $\mathcal{K}$ and $\mathcal{V}$ are $Z^T_{V}$.

% Then we use the gated fusion mechanism to fuse vision features and language representation. Gated fusion mechanism involves two steps: obtaining a score vector between 0 and 1 to determine the importance of each attention feature(Equation \ref{equ:fusion_sigmoid}) and using score to fuse text and attention features(Equation \ref{equ:fusion_lamda}).
As the roles of image features and text features in generating rationales and answers are not static or fixed, we choose to use a gate mechanism to fuse vision features and language representation, i.e., let the model decide how to use the image and text features. The gated fusion mechanism (~\cite{Zhang2020Neural}~\cite{wu2021good}~\cite{DBLP:conf/acl/LiLZZXMZ22}) involves two steps: obtaining a score vector between 0 and 1 to determine the importance of each feature (Equation \ref{equ:fusion_sigmoid}), and using the scores to fuse the text and attention features (Equation \ref{equ:fusion_lamda}).
\begin{equation}
\alpha = Sigmoid(W_l Z_{L} + W_v Z_{V}^{attn}),
\label{equ:fusion_sigmoid}
\end{equation}
\begin{equation}
Z_{fuse} = (1-\alpha) * Z_{L} + \alpha * Z_{V}^{attn}
\label{equ:fusion_lamda}
\end{equation}
where $W_l$ and $W_v$ are learnable parameters for gate projection, $Z_{fuse} = Z^R_{fuse}$ in the stage of rationale generation and $Z_{fuse} = Z^A_{fuse}$ in the stage of answer inference.

\subsubsection{Text Decoder}

%Finally, we put the output $Z_{fuse}$ into decoder of Transformer. The decoder generates rationale or answer in the stage of rationale generation and answer inference respectively. The algorithm is as Algorithm \ref{alg:method}.
In multi-modal CoT, the text decoder is responsible for generating rationales or inferring the final answers, taking into account the representation output $Z_{fuse}$ of the encoder and the previously decoded token to predict the next one. For example, in the stage of rationale generation, the decoder predicts the rationale $Y = (y_1, \dots , y_N)$ token by token, according to the last decoding state and source context. The rationale probability can be formulated as follows:
\begin{align}
s_i &= \textsc{SelfAttn}(Y_{<i}),\\
P(y_i|Y_{<i}, Z_{fuse}; \theta) &= Softmax(FFN(s_i + \nonumber\\
& \quad\quad \textsc{CrossAttn}(s_i, Z_{fuse}))),
\end{align}
where $\theta$ is the model parameters, $y_i$ is the $i$-th token in $Y$ with $N$ tokens, $s_i$ denotes the decoding state at the $i$-th timestep.

Therefore, the sequence generation loss $\mathcal{L}_{SEQ}$ for model optimization can be written as:
\begin{equation}
\mathcal{L}_{SEQ} = \sum_{i=1}^N -\log P(y_i|Y_{<i}, Z_{fuse}; \theta)
\label{equ:loss 1}
\end{equation}
where $Y = Y_R$ in the stage of rationale generation, and $Y = Y_A$ in the stage of answer inference.

% Latent Space Learning (多模态表征学习)
\subsection{Multi-modal Latent Space Learning}

% Our model takes input of images in PNG format. We utilize the idea of diffusion process to obtain better image features aligned with text representation. Firstly, we use the encoder of Variational AutoEncoder to get latent vector of image(Formula \ref{equ:image_encoder}). Then we add random noise to the latent vector which follows a Gaussian distribution with time steps. Next the latent vector will be put into UNet neural network. We fuse text representation and image features in a deep level by mapping text information into the intermediate layer of UNet through a cross attention layer. The noise predicted by UNet will compare with true noise. By this way, we train the model to get latent space of image. To utilize the attention mechanism between text representation and vision features, we leverage a linear layer $W_h$ to make dimension of vectors consistent.

% All in all, we adopt transformer model to encoder text input. Due to outstanding performance of diffusion model on getting latent space which can generate image features aligned with text representation, we adopt VAE and Unet model. By adding noise on image features gradually and predicting noise in accordance with time steps, we can obtain appropriate image features.

In the current multi-modal CoT work, such as MM-CoT~\cite{DBLP:journals/corr/abs-2302-00923}, image feature extraction is performed using off-the-shelf image encoders trained on models such as DETR~\cite{DBLP:conf/eccv/CarionMSUKZ20} and CLIP~\cite{DBLP:conf/icml/RadfordKHRGASAM21}. However, this method has two major drawbacks. Firstly, due to the limitations of pre-training objectives, the extracted image features are usually shallow and generic information that is not specifically optimized for reasoning, thus lacking deep semantic information which is required in reasoning. Secondly, the image features used for inference are highly dependent on language input, meaning that different image features are required for different language descriptions. In other words, the image features extracted independently and text features are not aligned. Therefore, this work proposes a method of multi-modal latent space learning, which learns flexible image features that are aligned with text inputs in the latent space and optimized for the inference process, thus possessing the deep semantics required for reasoning.

As Richard Feynman once said, "\textit{What I cannot create, I do not understand}." Therefore, we argue that excellent creativity must contain excellent understanding. Drawing inspiration from the outstanding generative performance of diffusion models, we apply the idea of a stable diffusion model to obtain a multi-modal latent space.
Specifically, we use the concept of a diffusion process to obtain better image features with deep semantics that align with text representation. Firstly, we employ the Variational AutoEncoder (VAE) \cite{DBLP:journals/corr/KingmaW13} as the image encoder to obtain the latent vector of the image. Then, we add random noise to the latent vector, which follows a Gaussian distribution with time steps. Next, the latent vector is inputted into the UNet neural network~\cite{DBLP:conf/miccai/RonnebergerFB15}. We fuse text representation and image features at a deep level by mapping text information into the intermediate layer of UNet through a cross-attention layer.
By optimizing the diffusion process, which compares the predicted noise with true noise, the model can obtain better image features with deep semantics that align with text inputs. This is because the diffusion process enables the model to learn features that are not only optimized for reasoning but also possess a high degree of stability and robustness to noise and other disturbances. In this way, the model can obtain the latent space of an image with deep semantics from the perspective of diffusion.

% \begin{algorithm}
%     \caption{Multi-modal Latent Space Learning for CoT}
%     \begin{algorithmic}[1] %每行显示行号
%     \REQUIRE Language Context $X_{L}$ Question $X_Q$, Options $X_O$, Image Input $X_{V}$
%     \ENSURE Rationale $\hat{Y_R}$, Answer $\hat{Y_A}$
%     \STATE Text Representation: \\
%     $Z_L^R \gets \textsc{Encoder}_{text}([X_L; X_Q; X_O])$
%     \STATE Multi-modal Latent Vector: \\
%     $Z_{V}^R \gets \textsc{Diffusion}(X_{V}, Z_L^R)$
%     \STATE Feature Fusion: \\
%     $Z_{fuse}^R \gets \textsc{Fuse}(Z_{V}^R, Z_L^R)$
%     \STATE Rationale Generation: \\
%     $\hat{Y_R} \gets \textsc{Decoder}_{text}(Z_{fuse}^R)$
%     \STATE Text Representation: \\
%     $Z_L^A \gets \textsc{Encoder}_{text}([X_L; X_Q; X_O; \hat{Y_R}])$
%     \STATE Multi-modal Latent Vector: \\
%     $Z_{V}^A \gets \textsc{Diffusion}(X_{V}, Z_L^A)$
%     \STATE Feature Fusion: \\
%     $Z_{fuse}^A \gets \textsc{Fuse}(Z_{V}^A, Z_L^A)$
%     \STATE Answer Inference: \\
%     $\hat{Y_A} \gets \textsc{Decoder}_{text}(Z_{fuse}^A)$
%     \end{algorithmic}
%     \label{alg:method}
% \end{algorithm}

Stable diffusion consists of two main parts: (1) the forward (or diffusion) process and the reverse process. In the diffusion process, random noise following a Gaussian distribution is added to the image features of latent space. This process is entirely run on the latent space and is composed of a VAE neural network and a scheduling algorithm. (2) The reverse process generates an image using an image decoder based on the latent features and text representations. Existing studies~\cite{DBLP:journals/corr/abs-2210-10960} have shown that the latent space already contains aligned semantic information, so we suppose that it can be utilized to fuse linguistic modality and visual modality for reasoning.

In our DPMM-CoT, we first encode the image into the latent space $Z_V^0$ using VAE. 
% \textcolor{red}{Especially, during inference, the image is encoded as a latent vector through the VAE and then directly fused with the text representation vector to generate a rational or answer.}
Especially, during inference, the image is encoded as a latent vector through the VAE and then directly fused with the text representation vector to generate a rational or answer.
Secondly, we add random noise that follows a Gaussian distribution to the latent space of the image.
\begin{align}
    Z_V^0 &= \eta VAE(X_V),\\
    q(Z_V^{t}|Z_V^{t-1}) &= \mathcal{N}(Z_V^{t};\sqrt{1-{\beta_{t}}}Z_V^{t-1},{\beta_{t}}\textbf{I})
    \label{equ:add_noise_by_step}
\end{align}
which indicates the diffusion process that adding noise that follows a Gaussian distribution, where ${\beta_t}$ is the variance schedule, ${\sqrt{1-{\beta_{t}}}Z_V^{t-1}}$ is the mean, $\textbf{I}$ is identity matrix, and $\eta = 0.18215$ is the scale factor. The analysis on value of $\eta$ is shown in Appendix.
%F \ref{sec:scale factor}.
%TODO: I 是什么
The diffusion process of the diffusion model can be expressed as a Markov chain from $t=0$ to $t=T$:
\begin{equation}
    q(Z_V^{0:T}) = q(Z_V^0)\prod_{t=1}^{T}q(Z_V^t|Z_V^{t-1}).
    \label{equ:add_noise_directly}
\end{equation}

\begin{table*}[t]
    \centering
    \small
    \caption{Main results on ScienceQA test set (\%). Size = backbone model size. Question classes: NAT = natural science, SOC = social science, LAN = language science, TXT = text context, IMG = image context, NO = no context, G1-6 = grades 1-6, G7-12 = grades 7-12. Results except ours are taken from \cite{DBLP:journals/corr/abs-2209-09513} and \cite{DBLP:journals/corr/abs-2302-00923}. Results in \textbf{bold} are the best performance.}\label{tab:main_results}
    % \scriptsize % 7 7
    % \footnotesize % 8 8
    % \small % 9 9
    % \normalsize % 10 10
    % \fontsize{8.3pt}{\baselineskip}\selectfont % font size
    \renewcommand\tabcolsep{5.5pt} % column space
    % \renewcommand\arraystretch{0.93} % line space
    % \resizebox{1.0\linewidth}{!}
    \vspace{-2mm}
    \begin{tabular}{l|r|cccccccc|l} 
    \toprule
     \bf Model  & Size & NAT & SOC & LAN & TXT & IMG & NO & G1-6 & G7-12 & ~Avg \\
    \midrule
    Human & - & 90.23  & 84.97 & 87.48 & 89.60 & 87.50 & 88.10 & 91.59 & 82.42 & 88.40 \\
     \midrule
     % Q only \citep{Anderson2017up} & FT & Q$\rightarrow$A & 41.34 & 27.22 & 47.00 & 41.79 & 35.15 & \fix{44.60} & 39.28 & 40.87  & 39.85 \\
     % C$_I$ only \citep{Anderson2017up} & FT & C$_I$$\rightarrow$A & 41.34 & 29.25 & 45.45 & 42.33 & 36.09 & \fix{42.93} & 39.21 & 41.07 & 39.87 \\
     % Q+M only \citep{Anderson2017up} & FT & QM$\rightarrow$A & 52.66 & 51.86 & 60.18 & 55.57 & 50.37 & \fix{57.42} & 52.53 & 57.88 & 54.44 \\
     % Q+C$_T$+M only \citep{Anderson2017up} & FT & QC$_T$M$\rightarrow$A & 57.28 & 49.04 & {61.36} & 60.46 & 52.80 & \fix{{58.82}} & 54.44 & 60.51 & 56.61 \\
     % Q+C$_I$+M only \citep{Anderson2017up} & FT & QC$_I$M$\rightarrow$A & {58.97} & {53.77} & 60.45 & {62.85} & {54.49} & \fix{57.63} & {56.72} & {61.04} & {58.26} \\
     % \midrule
     MCAN \cite{yu2019deep} & 95M & 56.08 & 46.23 & 58.09 & 59.43 & 51.17 & 55.40 & 51.65 & 59.72 & 54.54 \\
     Top-Down \cite{anderson2018bottom} & 70M & 59.50 & 54.33 & 61.82 & 62.90 & 54.88 & 59.79 & 57.27 & 62.16 & 59.02 \\
     BAN \cite{kim2018bilinear} & 112M & 60.88 & 46.57 & 66.64 & 62.61 & 52.60 & {65.51} & 56.83 & 63.94 & 59.37 \\
     DFAF \cite{gao2019dynamic} & 74M & 64.03 & 48.82 & 63.55 & 65.88 & 54.49 & 64.11 & 57.12 & 67.17 & 60.72 \\
     ViLT \cite{kim2021vilt} & 113M & 60.48 & 63.89 & 60.27 & 63.20 & 61.38 & 57.00 & 60.72 & 61.90 & 61.14 \\
     Patch-TRM \cite{lu2021iconqa} & 90M & {65.19} & 46.79 & {65.55} & {66.96} & 55.28 & 64.95 & 58.04 & {67.50} & 61.42 \\
     VisualBERT \cite{li2019visualbert} & 111M & 59.33 & {69.18} & 61.18 & 62.71 & {62.17} & 58.54 & {62.96} & 59.92 & {61.87} \\
     \midrule
     % UnifiedQA$_\text{SMALL}$ \citep{raffel2020exploring} & zero-shot& QCM$\rightarrow$A & 47.78 & 40.49 & 46.00 & 50.24 & 44.12 & \fix{44.39} & 45.56 & 46.21 & 45.79 \\
     % UnifiedQA$_\text{Base}$ \citep{raffel2020exploring} & zero-shot& QCM$\rightarrow$A & 50.13 & 44.54 & 48.18 & 53.08 & 48.09 & \fix{46.69} & 47.58 & 50.03 & 48.46 \\
      % UnifiedQA$_\text{SMALL}$ \citep{raffel2020exploring} & train set & QCM$\rightarrow$A & 53.77 & 58.04 & 61.09 & 52.10 & 51.51 & \fix{61.46} & 58.22 & 53.59 & 56.57 \\
     UnifiedQA$_\textrm{Base}$ \cite{khashabi-etal-2020-unifiedqa}& 223M &68.16 & 69.18 & 74.91 & 63.78 & 61.38 & 77.84 & 72.98 & 65.00 & 70.12 \\
     % \textbf{UnifiedQA$_\text{Base}$ (CoT)} & train set & QCM$\rightarrow$AE & 70.60 & 74.02 & 78.36 & 65.69 & 64.80 & \fix{81.53} & 75.48 & {69.48} & 73.33$_{3.21 \uparrow}$ \\
     UnifiedQA$_\textrm{Base}$ w/ CoT \cite{DBLP:journals/corr/abs-2209-09513} & 223M & {71.00} & {76.04} & {78.91} & {66.42} & {66.53} & {81.81} & {77.06} & 68.82 & {74.11}  \\
     \midrule
     ChatGPT (GPT-3.5) & 175B & 74.64 & 69.74 & 76.00 & 74.44 & 67.28 & 77.42 & 76.80 & 68.89 & 73.97 \\
    ChatGPT (GPT-3.5) w/ CoT \cite{DBLP:journals/corr/abs-2209-09513} & 175B& {75.44} & {70.87} & {78.09} & {74.68} & {67.43} & {{79.93}} & {78.23} & {69.68} & {75.17} \\
     \midrule
     MM-CoT$_\textrm{Base}$(CLIP) & 223M+151M  & {87.97} & {80.88} & {87.36} & {88.32} & {84.78} & {88.15} & {86.34} & {86.29} & {86.32} \\
     \midrule
     MM-CoT$_\textrm{Base}$(DETR) \cite{DBLP:journals/corr/abs-2302-00923}& 223M+60M  & {87.52} & {77.17} & {85.82} & {87.88} & {82.90} & {86.83} & {84.65} & {85.37} & {84.91} \\
     \textbf{DPMM-CoT$_\textrm{Base}$} & 223M+83M  & \textbf{92.72} & \textbf{87.85} & \textbf{89.91} & \textbf{92.72} & \textbf{90.48} & \textbf{91.29} & \textbf{91.45} & \textbf{90.11} & \textbf{90.97} \\
     \midrule
     MM-CoT$_\textrm{Large}$(DETR) \cite{DBLP:journals/corr/abs-2302-00923}& 738M+60M & \textbf{95.91} & 82.00 & 90.82 & {95.26} & {88.80} & \textbf{92.89} & {92.44} & {90.31} & {91.68} \\
     \textbf{DPMM-CoT$_\textrm{Large}$} & 738M+83M & {95.52} & \textbf{90.33} & \textbf{91.36} & \textbf{95.50} & \textbf{93.26} & {{92.68}} & \textbf{93.28} & \textbf{93.47} & \textbf{93.35} \\
     \bottomrule
    \end{tabular}
%\vspace{-3mm}
\end{table*}

\begin{table}[htb]
    \vspace{-3mm}
    \centering\small
    \caption{Performance of rationale generation and answer inference. \label{tab:rationale}}
    \vspace{-3mm}
    \begin{tabular}{lcc}\toprule
     {Method} & {(i) QCM$ \rightarrow$ R}  & {(ii) QCMR$ \rightarrow$ A}  \\\midrule
     MM-CoT$_\textrm{Base}$ & 96.97 &  84.91 \\
     \midrule
     DPMM-CoT$_\textrm{Base}$ & 98.18 & 90.97 \\
    \bottomrule
    \end{tabular}
    \vspace{-3mm}
\end{table}

When $T\rightarrow{\infty}$, the final result will become a noisy image, similar to sampling from an isotropic Gaussian distribution. However, we also use a closed-form formula to directly sample noisy images at a specific time step $t$, instead of designing an algorithm to iteratively add noise to the image following the practice of \cite{DBLP:conf/cvpr/RombachBLEO22}.

%\textcolor{red}{(Omitting formula derivation)}
\begin{equation}
\begin{aligned}
\setlength{\abovedisplayskip}{3pt}
\setlength{\belowdisplayskip}{3pt}
 Z_V^t
 % &= \sqrt{1-\beta_t}Z_V^{t-1} + \sqrt{\beta_t}\epsilon_{t-1}\\
% &=\sqrt{\alpha_t}Z_V^{t-1} + \sqrt{1-\alpha_t}\epsilon_{t-1}\\
% &=\sqrt{\alpha_t}(\sqrt{\alpha_{t-1}}Z_V^{t-2} + \sqrt{1-\alpha_{t-1}}\epsilon_{t-2}) + \sqrt{1-\alpha_t}\epsilon_{t-1}\\
% &=\sqrt{\alpha_t\alpha_{t-1}}Z_V^{t-2} + \sqrt{\alpha_t(1-\alpha_{t-1})}\epsilon_{t-2} + \sqrt{1-\alpha_t}\epsilon_{t-1}\\
% &=\sqrt{\alpha_t\alpha_{t-1}}Z_V^{t-2} + \sqrt{1-\alpha_t\alpha_{t-1}}\overline\epsilon_{t-2}\\
% &\vdots\\
% &=\sqrt{\alpha_t\alpha_{t-1}\cdots\alpha_1}Z_V^0 + \sqrt{1-\alpha_t\alpha_{t-1}\cdots\alpha_1}\epsilon\\
&=\sqrt{\overline{\alpha}_t}Z_V^0 + \sqrt{1-\overline{\alpha}_t}\epsilon
\end{aligned}
\label{equ:diffusion_deduce}
\end{equation}
where $\alpha_t$ = 1 - $\beta_t$, $\overline{\alpha}_t$ = $\prod_{i=1}^t\alpha_i$. $\epsilon$ is an i.i.d. (independent identically distributed) standard normal random variable. It is important to distinguish them using different symbols and subscripts because they are independent and their values may differ after sampling. 

The standard diffusion process involves predicting the noise using UNet~\cite{DBLP:conf/miccai/RonnebergerFB15}. By utilizing a cross-attention layer to map text information into the intermediate layer of UNet, we can merge text representation with image features. This integration of information from both modalities leads to a more comprehensive understanding of the underlying structure in the data.
Text features provide valuable additional semantic information that may not be immediately evident from the visual content alone. Incorporating these features into the model allows us to better comprehend the context and meaning behind the visual elements.
Meanwhile, image features offer rich visual information about the objects and scenes depicted in the image. These features enable the model to identify patterns and relationships between different parts of the image.
The fusion of both image and text features through diffusion process enables the model to leverage the strengths of both modalities, leading to improved latent space learning. 

Specifically, we predict the noise $\epsilon_{\theta}(Z_V^t, t, Z_L)$ by UNet with the attention mechanism between visual feature $Z_V^t$ and text representation $Z_L$ as follows:
\begin{equation}
    \epsilon_{\theta}(Z_V^t, t, Z_L) = \textsc{UNet}(FFN(Softmax(\frac{Q K^T}{\sqrt{d}}) V + Q))
    \label{equ_attention_in_UNet}
\end{equation}
where 
$Q = W_Q^{(i)}\cdot Z_V^t$, 
$K = W_K^{(i)}\cdot Z_L$, 
$V = W_V^{(i)}\cdot Z_L$. 
$Z_V^t \in \mathbb{R} ^{N \times d^i}$ 
is an intermediate representation of UNet. 
Therefore, the latent diffusion process loss implemented with Maximum Square Error (MSE) can be written as:
%, and $W_Q^{(i)}$, $W_K^{(i)}$ and $W_V^{(i)}$ are learnable parameters
\begin{equation}
\mathcal{L}_{LDM} = \mathbb{E}_{\epsilon\sim\mathcal{N}(0.1), Z_L, t}[||\epsilon-\epsilon_{\theta}(Z_V^t, t, Z_L)||_2^2]
\label{equ:loss 2}
\end{equation}
where $\mathcal{L}_{LDM}$ is the loss of latent diffusion model, $\theta$ is the parameters of the model, $\epsilon$ is the random noise (an independent identically distributed standard normal random variable). 
So the total loss of the model is as follows:
\begin{equation}
\mathcal{L}_{total} = \mathcal{L}_{SEQ} + \mathcal{L}_{LDM}.
\end{equation}
% \textcolor{red}{
% During the training process, all model parameters, except for the UNet parameters in the reverse process, are updated. The VAE parameters and Text encoder parameters are updated based on both $\mathcal{L}_{SEQ}$ and $\mathcal{L}_{LDM}$, while the Feature fusion module and Text decoder are solely updated through $\mathcal{L}_{SEQ}$.
% }
During the training process, all model parameters, except for the UNet parameters in the reverse process, are updated. The VAE parameters and Text encoder parameters are updated based on both $\mathcal{L}_{SEQ}$ and $\mathcal{L}_{LDM}$, while the Feature fusion module and Text decoder are solely updated through $\mathcal{L}_{SEQ}$.

\section{Experiments}

\subsection{Setup}

\subsubsection{Datasets}

To assess CoT on LLMs, we followed the approach of MM-CoT~\cite{DBLP:journals/corr/abs-2302-00923} and used the Science Question Answering (ScienceQA)~\cite{DBLP:journals/corr/abs-2209-09513} dataset. 

% This dataset comprises approximately 21,000 multi-modal multiple-choice questions from various scientific fields. Each example includes a question, potential answers, a correct answer, context, and a series of explanations and solutions. The dataset covers three subjects, 26 topics, 127 categories, and 379 skills. It is divided into training, validation, and test sets, containing 12,726, 4,241, and 4,241 examples, respectively. Note that some images in the dataset come with captions; however, we chose not to use them in our model since their inclusion for image replacement can result in significant information loss.

In addition, to demonstrate the general effectiveness of the Diffusion Process for Multi-modal Latent Space Learning, we also conducted experiments on a multi-modal machine translation task. We selected the Multi30K multi-modal translation dataset~\cite{elliott-etal-2016-multi30k} and followed the work of IKD-MMT~\cite{DBLP:conf/emnlp/PengZZ22}. 
% This dataset extends the
% Flickr30K dataset~\cite{DBLP:journals/ijcv/PlummerWCCHL17} which consists of 30K images and 150K descriptive captions. And Multi30K comprises translations created by professional translators over a subset of the English descriptions. We conducted experiments on English-to-German and English-to-French translations, using three test sets: test2016-flickr (Test16), test2017-flickr (Test17), and test2017-mscoco (MSCOCO) containing 1000, 1000, 461 examples, respectively.
For a detailed introduction to the above dataset, please refer to the Appendix.

\begin{table}[t]
    \small
    \setlength\tabcolsep{2pt}
    \centering
    \caption{BLEU score of EN-DE and EN-FR tasks. Encouragingly, our DPMM-MT is an image-must MMT model.\\
    % Transformer\cite{vaswani2017attention};
    % Fusion-conv\cite{caglayan-etal-2017-lium};
    % Trg-mul\cite{caglayan-etal-2017-lium};
    % UVR-NMT\cite{Zhang2020Neural};
    % GMNMT\cite{yin-etal-2020-novel};
    % DCCN\cite{lin2020dynamic};
    % ImagiT \cite{long-etal-2021-generative};
    % Gated Fusion\cite{wu2021good};
    % RMMT\cite{wu2021good};
    IKD-MMT \cite{DBLP:conf/emnlp/PengZZ22}
    }
    \label{tab:translation}
    \vspace{-3mm}
    {\begin{tabular}{lccccc}
    \toprule
    \multicolumn{1}{c}{\multirow{2}*{\textbf{Model}}} & \multicolumn{3}{c}{\textbf{EN-DE}} & \multicolumn{2}{c}{\textbf{EN-FR}} \\
    \cmidrule(lr){2-4} \cmidrule(lr){5-6}
    & \multicolumn{1}{c}{\textbf{Test16}} & \multicolumn{1}{c}{\textbf{Test17}} & \multicolumn{1}{c}{\textbf{MSCOCO}} & \multicolumn{1}{c}{\textbf{Test16}} & \multicolumn{1}{c}{\textbf{Test17}} \\
    \midrule
    % \multicolumn{1}{c}{\textit{Language-only MMT}}\\
    % Transformer & 37.6 & 31.7 & 27.9 & 59.0 & 51.9  \\
    % Multitask\cite{elliott-kadar-2017-imagination} & 36.8 & - & - & - & -  \\
    % NMT$_{\text{SRC+IMG}}$\cite{DBLP:conf/acl/CalixtoLC17} & 36.5 & - & - & - & - \\
    % IMG$_D$\cite{DBLP:conf/emnlp/CalixtoL17} & 37.3 & - & - & - & -\\
    % Fusion-conv & 37.0 & 29.8 & 25.1 & 53.5 & 51.6\\
    % Trg-mul & 37.8 & 30.7 & 26.4 & 54.7 & 52.7\\
    % VAG-NMT\cite{zhou-etal-2018-visual} & - & 31.6 & 28.3 & - & 53.8 \\
    % VMMT$_\text{F}$\cite{calixto-etal-2019-latent} & 37.7 & 30.1 & 25.5 & - & - \\
    % DS-SUM-L2\cite{caglayan2019multimodal} & 39.4 & 32.6 & - & 60.7 & 54.2 \\
    % Del+obj\cite{ive-etal-2019-distilling} & 38.0 & - & - & 59.8 & - \\
    % UVR-NMT & 36.94 & 28.63 & - & 57.53 & 48.46 \\
    % Multimodal\cite{yao-wan-2020-multimodal} & 38.7 & - & - & - & - \\
    % GMNMT & 39.8 & 32.2 & 28.7 & 60.9 & 53.9 \\
    % DCCN & 39.7 & 31.0 & 26.7 & 61.2 & 54.3 \\
    % ImagiT & 38.5 & 32.1 & 28.7 & 59.7 & 52.4 \\

    % \midrule
    % \multicolumn{1}{c}{\textit{Multi-modal MMT}}\\
    
    % Gumbel-att\cite{liu2021gumbel} & 39.2 & 31.4 & 26.9 & - & - \\
    % OVC+${L}_{m}$\cite{wang2021efficient} & - & 32.3 & 28.9 & - & 54.1 \\
    % Gated Fusion & \textbf{41.96} & 33.59 & 29.04 & 61.69 & 54.85 \\
    % RMMT & 41.45 & 32.94 & 30.01 & 62.1 & 54.39 \\
    IKD-MMT  & 41.28 & 33.83 & 30.17 & 62.53 & 54.84 \\
    \hdashline
    {mT5} & 38.56 & {33.01} & {28.10} & {61.71} & {53.84} \\
    \textbf{DPMM-MT} & 41.63 & \textbf{36.18} & \textbf{30.75} & \textbf{66.91} & \textbf{57.80} \\
    \bottomrule
\end{tabular}}
\end{table}

\begin{table*}[t]
\centering
\small
\caption{Results of different way of solving problems without images.}
\vspace{-3mm}
\renewcommand\tabcolsep{9.8pt} % column space
{
\begin{tabular}{l|cccccccc|c}
\toprule
 Model  & NAT & SOC & LAN & TXT & IMG & NO & G1-6 & G7-12 & ~Avg \\
 \midrule
Zero Tensor & {92.72} & {87.85} & {89.91} & {92.72} & {90.48} & {91.29} & {91.45} & {90.11} & {90.97} \\
Blank Image & {92.54} & {82.56} & {89.91} & {92.86} & {88.35} & {90.94} & {90.68} & {88.13} & {89.77} \\
% \quad w/o both & 81.04 & 77.17 & 81.55 & 79.96 & 74.22 & 83.97 & 80.36 & 80.36 & 80.36 \\
 \bottomrule
\end{tabular}
}
 \label{tab:void_image}
 \vspace{-3mm}
\end{table*}

\subsubsection{Settings} 

In our experiment on CoT in LLMs, we employed a two-stage framework consisting of two procedures: rationale generation and answer inference. Both stages shared the same model architecture, namely the T5 encoder-decoder architecture~\cite{DBLP:journals/jmlr/RaffelSRLNMZLL20}.

For our experiment on multi-modal machine translation (MMT), we employed the mT5 encoder-decoder architecture, which was initialized using the pre-trained mT5-large~\cite{DBLP:conf/naacl/XueCRKASBR21} checkpoint, which had been pre-trained on a multilingual corpus.
Please refer to the Appendix for detailed settings.

\subsection{Main Analysis}

Table \ref{tab:main_results} presents the main results of our study, which compares the performance of various Visual Question Answering (VQA) models. We evaluated our DPMM-CoT model against MM-CoT, a baseline, and found that DPMM-CoT$_\textrm{Base}$ outperforms MM-CoT$_\textrm{Base}$(DETR) by 6.06\% and DPMM-CoT$_\textrm{Large}$ outperforms MM-CoT$_\textrm{Large}$(DETR) by 1.67\%. 
Notably, when questions involve visual context (IMG column), DPMM-CoT$_\textrm{Base}$ and DPMM-CoT$_\textrm{Large}$ outperform MM-CoT$_{Base}$(DETR) and MM-CoT$_\textrm{Large}$(DETR) by 7.58\% and 4.46\%, respectively.

Compared to other VQA baselines, DPMM-CoT$_\textrm{Large}$ outperforms VisualBERT~\cite{li2019visualbert} by 31.48\%, demonstrating that autoregressive language pre-training and larger language models are effective for problem solving. And DPMM-CoT$_\textrm{Large}$ surpasses the UnifiedQA model with CoT~\cite{DBLP:journals/corr/abs-2209-09513} by 19.24\%. This suggests that only leveraging captions of images as visual context causes severe information loss and hallucination in CoT. 

Additionally, we found that DPMM-CoT$_\textrm{Large}$ outperforms the strong LLM -- ChatGPT by 18.18\%, demonstrating that language models under 1B parameters can perform better than general LLMs when fine-tuned with appropriate network designs and information. Moreover, our DPMM-CoT$_\textrm{Base}$ and DPMM-CoT$_\textrm{Large}$ both outperform human performance, indicating the effectiveness of our model. These results suggest that multi-modal latent space learning is significant for understanding flexible and deep visual information. In Table \ref{tab:rationale}, the ROUGE-L results of rationals generated by DPMM-CoT$_\textrm{Base}$ and MM-CoT$_\textrm{Base}$, as well as the accuracy of answers inferred under a two-stage framework, are shown. The specific analysis is shown in Appendix.%G\ref{sec:quality of rationale}.

To verify that the improvement of DPMM-CoT$_{\textrm{Large}}$ originates from multi-modal latent space learning via the diffusion process rather than an increase in the number of parameters, we utilized fixed visual features extracted by clip-vit-base-patch32~\cite{DBLP:journals/corr/abs-2302-00923}, which has 151M parameters. The result shows that while an increase in the number of parameters may contribute to improved performance on multi-modal QA tasks, it is still far from our DPMM-CoT model. This suggests that our improvements are due to a deeper understanding of visual information gained through multi-modal latent space learning. The case study is shown in Appendix.

\begin{table*}[t]
\centering\small
% \vspace{-1mm}
\caption{Ablation results of our method.}
% \scriptsize % 7 7
% \footnotesize % 8 8
% \small % 9 9
% \normalsize % 10 10
% \fontsize{8.3pt}{\baselineskip}\selectfont % font size
\renewcommand\tabcolsep{9.8pt} % column space
% \renewcommand\arraystretch{0.93} % line space
% \resizebox{1.0\linewidth}{!}
\vspace{-3mm}
{
\begin{tabular}{l|cccccccc|c}
\toprule
 Model  & NAT & SOC & LAN & TXT & IMG & NO & G1-6 & G7-12 & ~Avg \\
 \midrule
Our model & {92.72} & {87.85} & {89.91} & {92.72} & {90.48} & {91.29} & {91.45} & {90.11} & {90.97} \\
\quad w/o Stable Diffusion Pre-training & 88.63 & 80.43 & 85.45 & 89.93 & 84.88 & 85.92 & 87.15 & 84.18 & 86.09 \\
\quad w/o UNet & 91.92 & 82.56 & 89.91 & 92.18 & 88.35 & 90.52 & 89.98 & 88.46 & 89.44 \\
\quad w/ Frozen VAE & 91.07 & 82.00 & 90.36 & 91.64 & 87.36 & 90.73 & 89.35 & 88.33 & 88.99 \\
%\quad w/o Updating VAE & 92.76 & 87.40 & 89.64 & 92.77 & 90.18 & 91.15 & 91.59 & 89.45 & 90.83 \\
%\quad w/o UNet & 92.81 & 88.53 & 90.18 & 92.86 & 90.43 & 91.36 & 91.92 & 89.98 & 91.23 \\
% \quad w/o both & 81.04 & 77.17 & 81.55 & 79.96 & 74.22 & 83.97 & 80.36 & 80.36 & 80.36 \\
 \bottomrule
\end{tabular}
}
 \label{tab:ablation_results}
 \vspace{-3mm}
\end{table*}

\subsection{Further Analysis}

\subsubsection{Generalization to More Multi-modal Tasks}

To demonstrate the generality of our method across different multimodal tasks, we conducted experiments on Multimodal Machine Translation (MMT). The main results are presented in Table \ref{tab:translation}. We trained our Diffusion Process Enhanced Multi-Modal Machine Translation (DPMM-MT) model on the Multi30K dataset, which includes English-to-French and English-to-German translations. We then evaluated DPMM-MT on three test sets: test2016-flickr (Test16), test2017-flickr (Test17), and test2017-mscoco (MSCOCO).
Firstly, compared to the mT5 baseline that does not use image features, we achieved significant improvements in En-De of 3.07, 3.17 and 2.65, and in En-Fr of 5.20 and 3.96, respectively. This indicates the crucial role of image features in multimodal machine translation.
We achieved a new state-of-the-art (SOTA) result on Test17 and MSCOCO with English-to-German translation and Test16 and Test17 with English-to-French translation. Specifically, DPMM-MT outperformed the previous SOTA by 2.35 (33.83 $\rightarrow$ 36.18) and 0.58 (30.17 $\rightarrow$ 30.75) on Test17 and MSCOCO with English-to-German translation, respectively. DPMM-MT outperformed the previous SOTA by 4.38 (62.53 $\rightarrow$ 66.91) and 2.96 (54.84 $\rightarrow$ 57.80) on Test16 and Test17 with English-to-French translation, respectively. For Test16 with English-to-German translation, we also achieved comparable results to the previous SOTA - Gated Fusion \cite{wu2021good}.
These improvements across multiple datasets suggest that utilizing our proposed multi-modal latent space learning to extract deep image semantics is useful for enhancing the performance of multi-modal machine translation.

\subsubsection{Problems without Images}

Since not all questions in the ScienceQA task (or other real-life tasks) include images, our method needed to be adaptable to image-less questions. For this purpose, we explored two approaches: using blank images or null tensors as input for these questions. We analyzed the results using models DPMM-CoT$_{\textrm{Base}}$, and the experiment outcomes are presented in Table \ref{tab:void_image}. Our findings show that using zero tensors resulted in a 1.20\% higher accuracy than using blank images. This may be attributed that blank images may introduce misleading information during the diffusion process.

% \subsubsection{Stability of Improvements}

% To better demonstrate the confidence level of our method's performance compared to MM-CoT, we visualized the evaluation accuracy curve of both models in different training epochs under the same conditions. Figure \ref{fig:epoch} showcases our proposed multi-modal latent space learning's ability to bring stable benefits throughout the entire training process, rather than just specific moments, as evidenced by the improved performance in all epochs compared to the baseline.

\subsection{Ablation Study}

To illustrate the effect of each component in the Diffusion Process on multi-modal latent space learning, we conducted an ablation study. 
As shown in Table \ref{tab:ablation_results}, we tested whether pre-trained stable diffusion module is useful for multi-modal latent space learning. We randomly initialized the parameters of UNet and VAE, and evaluated the result without Stable Diffusion Pre-training. The results show that diffusion models including VAE and UNet initialized from pre-trained model are indeed useful for DPMM-CoT. The accuracy declined by 4.88\% (90.97\% $\rightarrow$ 86.09\%), demonstrating the importance of good initialization for producing effective multi-modal latent space.
Furthermore, we found that diffusion components initialized by random parameters actually outperform the baseline MM-CoT(DETR). This highlights the ability of the diffusion process to deeply understand image information after being trained on the ScienceQA dataset, producing effective image features aligned with text representation.

To further demonstrate the importance of diffusion process in multi-modal latent space learning, we trained the model without UNet. The images were only encoded by VAE to produce latents. The result in Table \ref{tab:ablation_results} shows that accuracy declined by 1.53\% (90.97\% $\rightarrow$ 89.44\%), indicating the significance of diffusion process to produce multi-modal latent space.
% \textcolor{red}{to produce multi-modal latent space.} 
These results testify that diffusion process is a key part of multi-modal latent space learning, and visual feature extraction by encoder alone is insufficient. By adding noise and predicting noise with UNet guided by text representation, the multi-modal latent space learning gains a deep understanding of image with language thoughts.

The quality of the vision latent vector that VAE produces has a significant impact on the effectiveness of the CoT. To prove this, we tried not updating the parameters of VAE during CoT training but instead used pre-trained parameters from Stable-Diffusion-v1-4. The results (90.97\% $\rightarrow$ 88.99\%) show that VAE trained with CoT is helpful in producing better latent vectors for use in reasoning. This also demonstrates that for reasoning tasks, it's not enough to perform only self-supervised pre-training.

\section{Conclusion}

In this work,  we focuses on improving the production of multi-modal latent spaces that can effectively understand both linguistic and visual information at a deeper level. To achieve this, we introduce DPMM-CoT, a multi-modal latent space learning approach via diffusion process for CoT reasoning in language models. Our experimental results demonstrate that our method performs exceptionally well on multi-modal tasks. Notably, DPMM-CoT$_\textrm{Base}$ outperforms MM-CoT$_\textrm{Base}$ by 6.06\%, while DPMM-CoT$_\textrm{Large}$ outperforms MM-CoT$_\textrm{Large}$ by 1.67\%. We also conducted additional experiments on multi-modal machine translation, which verified the effectiveness of our proposed multi-modal latent space learning method on a wider range of multi-modal tasks.
Moreover, our concrete analysis shows that our method enables language models to attain deeper, more flexible, and aligned features for language thought, thereby enhancing their reasoning abilities. In the future, we plan to evaluate our method on more multi-modal tasks.

% \section{Acknowledgments}
% \bigskip
% \noindent Thank you for reading these instructions carefully. We look forward to receiving your electronic files!

\bibliography{aaai23}

\begin{thebibliography}{38}
\providecommand{\natexlab}[1]{#1}

\bibitem[{Anderson et~al.(2018)Anderson, He, Buehler, Teney, Johnson, Gould,
  and Zhang}]{anderson2018bottom}
Anderson, P.; He, X.; Buehler, C.; Teney, D.; Johnson, M.; Gould, S.; and
  Zhang, L. 2018.
\newblock Bottom-up and top-down attention for image captioning and visual
  question answering.
\newblock In \emph{Proceedings of the IEEE conference on computer vision and
  pattern recognition}, 6077--6086.

\bibitem[{Bird, Klein, and Loper(2009)}]{Bird_Natural_Language_Processing_2009}
Bird, S.; Klein, E.; and Loper, E. 2009.
\newblock \emph{{Natural Language Processing with Python: Analyzing Text with
  the Natural Language Toolkit}}.
\newblock O'Reilly Media, Inc.

\bibitem[{Caglayan et~al.(2017)Caglayan, Aransa, Bardet,
  Garc{\'\i}a-Mart{\'\i}nez, Bougares, Barrault, Masana, Herranz, and van~de
  Weijer}]{caglayan-etal-2017-lium}
Caglayan, O.; Aransa, W.; Bardet, A.; Garc{\'\i}a-Mart{\'\i}nez, M.; Bougares,
  F.; Barrault, L.; Masana, M.; Herranz, L.; and van~de Weijer, J. 2017.
\newblock {LIUM}-{CVC} Submissions for {WMT}17 Multimodal Translation Task.
\newblock In \emph{Proceedings of the Second Conference on Machine
  Translation}, 432--439.

\bibitem[{Carion et~al.(2020)Carion, Massa, Synnaeve, Usunier, Kirillov, and
  Zagoruyko}]{DBLP:conf/eccv/CarionMSUKZ20}
Carion, N.; Massa, F.; Synnaeve, G.; Usunier, N.; Kirillov, A.; and Zagoruyko,
  S. 2020.
\newblock End-to-End Object Detection with Transformers.
\newblock In Vedaldi, A.; Bischof, H.; Brox, T.; and Frahm, J., eds.,
  \emph{Computer Vision - {ECCV} 2020 - 16th European Conference, Glasgow, UK,
  August 23-28, 2020, Proceedings, Part {I}}, volume 12346 of \emph{Lecture
  Notes in Computer Science}, 213--229. Springer.

\bibitem[{Elliott et~al.(2016)Elliott, Frank, Sima{'}an, and
  Specia}]{elliott-etal-2016-multi30k}
Elliott, D.; Frank, S.; Sima{'}an, K.; and Specia, L. 2016.
\newblock {M}ulti30{K}: Multilingual {E}nglish-{G}erman Image Descriptions.
\newblock In \emph{Proceedings of the 5th Workshop on Vision and Language},
  70--74. Berlin, Germany: Association for Computational Linguistics.

\bibitem[{Gao et~al.(2019)Gao, Jiang, You, Lu, Hoi, Wang, and
  Li}]{gao2019dynamic}
Gao, P.; Jiang, Z.; You, H.; Lu, P.; Hoi, S.~C.; Wang, X.; and Li, H. 2019.
\newblock Dynamic fusion with intra-and inter-modality attention flow for
  visual question answering.
\newblock In \emph{Proceedings of the IEEE/CVF conference on computer vision
  and pattern recognition}, 6639--6648.

\bibitem[{Goodfellow et~al.(2014)Goodfellow, Pouget{-}Abadie, Mirza, Xu,
  Warde{-}Farley, Ozair, Courville, and
  Bengio}]{DBLP:conf/nips/GoodfellowPMXWOCB14}
Goodfellow, I.~J.; Pouget{-}Abadie, J.; Mirza, M.; Xu, B.; Warde{-}Farley, D.;
  Ozair, S.; Courville, A.~C.; and Bengio, Y. 2014.
\newblock Generative Adversarial Nets.
\newblock In Ghahramani, Z.; Welling, M.; Cortes, C.; Lawrence, N.~D.; and
  Weinberger, K.~Q., eds., \emph{Advances in Neural Information Processing
  Systems 27: Annual Conference on Neural Information Processing Systems 2014,
  December 8-13 2014, Montreal, Quebec, Canada}, 2672--2680.

\bibitem[{Ho, Jain, and Abbeel(2020)}]{DBLP:conf/nips/HoJA20}
Ho, J.; Jain, A.; and Abbeel, P. 2020.
\newblock Denoising Diffusion Probabilistic Models.
\newblock In Larochelle, H.; Ranzato, M.; Hadsell, R.; Balcan, M.; and Lin, H.,
  eds., \emph{Advances in Neural Information Processing Systems 33: Annual
  Conference on Neural Information Processing Systems 2020, NeurIPS 2020,
  December 6-12, 2020, virtual}.

\bibitem[{Khashabi et~al.(2020)Khashabi, Min, Khot, Sabharwal, Tafjord, Clark,
  and Hajishirzi}]{khashabi-etal-2020-unifiedqa}
Khashabi, D.; Min, S.; Khot, T.; Sabharwal, A.; Tafjord, O.; Clark, P.; and
  Hajishirzi, H. 2020.
\newblock {UNIFIEDQA}: Crossing Format Boundaries with a Single {QA} System.
\newblock In \emph{Findings of the Association for Computational Linguistics:
  EMNLP 2020}, 1896--1907. Online: Association for Computational Linguistics.

\bibitem[{Kim, Jun, and Zhang(2018)}]{kim2018bilinear}
Kim, J.-H.; Jun, J.; and Zhang, B.-T. 2018.
\newblock Bilinear attention networks.
\newblock \emph{Advances in neural information processing systems}, 31.

\bibitem[{Kim, Son, and Kim(2021)}]{kim2021vilt}
Kim, W.; Son, B.; and Kim, I. 2021.
\newblock Vilt: Vision-and-language transformer without convolution or region
  supervision.
\newblock In \emph{International Conference on Machine Learning}, 5583--5594.
  PMLR.

\bibitem[{Kingma and Welling(2014)}]{DBLP:journals/corr/KingmaW13}
Kingma, D.~P.; and Welling, M. 2014.
\newblock Auto-Encoding Variational Bayes.
\newblock In Bengio, Y.; and LeCun, Y., eds., \emph{2nd International
  Conference on Learning Representations, {ICLR} 2014, Banff, AB, Canada, April
  14-16, 2014, Conference Track Proceedings}.

\bibitem[{Kojima et~al.(2022)Kojima, Gu, Reid, Matsuo, and
  Iwasawa}]{DBLP:journals/corr/abs-2205-11916}
Kojima, T.; Gu, S.~S.; Reid, M.; Matsuo, Y.; and Iwasawa, Y. 2022.
\newblock Large Language Models are Zero-Shot Reasoners.
\newblock \emph{CoRR}, abs/2205.11916.

\bibitem[{Kwon, Jeong, and Uh(2022)}]{DBLP:journals/corr/abs-2210-10960}
Kwon, M.; Jeong, J.; and Uh, Y. 2022.
\newblock Diffusion Models already have a Semantic Latent Space.
\newblock \emph{CoRR}, abs/2210.10960.

\bibitem[{Li et~al.(2022)Li, Lv, Zhou, Zhou, Xiao, Ma, and
  Zhu}]{DBLP:conf/acl/LiLZZXMZ22}
Li, B.; Lv, C.; Zhou, Z.; Zhou, T.; Xiao, T.; Ma, A.; and Zhu, J. 2022.
\newblock On Vision Features in Multimodal Machine Translation.
\newblock In Muresan, S.; Nakov, P.; and Villavicencio, A., eds.,
  \emph{Proceedings of the 60th Annual Meeting of the Association for
  Computational Linguistics (Volume 1: Long Papers), {ACL} 2022, Dublin,
  Ireland, May 22-27, 2022}, 6327--6337. Association for Computational
  Linguistics.

\bibitem[{Li et~al.(2019)Li, Yatskar, Yin, Hsieh, and Chang}]{li2019visualbert}
Li, L.~H.; Yatskar, M.; Yin, D.; Hsieh, C.-J.; and Chang, K.-W. 2019.
\newblock Visualbert: A simple and performant baseline for vision and language.
\newblock \emph{arXiv preprint arXiv:1908.03557}.

\bibitem[{Lin(2003)}]{lin2003rouge}
Lin, C.-Y. 2003.
\newblock ROUGE: Recall-oriented understudy for gisting evaluation.

\bibitem[{Lin et~al.(2020)Lin, Meng, Su, Yin, Yang, Ge, Zhou, and
  Luo}]{lin2020dynamic}
Lin, H.; Meng, F.; Su, J.; Yin, Y.; Yang, Z.; Ge, Y.; Zhou, J.; and Luo, J.
  2020.
\newblock Dynamic context-guided capsule network for multimodal machine
  translation.
\newblock In \emph{Proceedings of the 28th ACM International Conference on
  Multimedia}, 1320--1329.

\bibitem[{Long, Wang, and Li(2021)}]{long-etal-2021-generative}
Long, Q.; Wang, M.; and Li, L. 2021.
\newblock Generative Imagination Elevates Machine Translation.
\newblock In \emph{Proceedings of the 2021 Conference of the North American
  Chapter of the Association for Computational Linguistics: Human Language
  Technologies}, 5738--5748.

\bibitem[{Lu et~al.(2022)Lu, Mishra, Xia, Qiu, Chang, Zhu, Tafjord, Clark, and
  Kalyan}]{DBLP:journals/corr/abs-2209-09513}
Lu, P.; Mishra, S.; Xia, T.; Qiu, L.; Chang, K.; Zhu, S.; Tafjord, O.; Clark,
  P.; and Kalyan, A. 2022.
\newblock Learn to Explain: Multimodal Reasoning via Thought Chains for Science
  Question Answering.
\newblock \emph{CoRR}, abs/2209.09513.

\bibitem[{Lu et~al.(2021)Lu, Qiu, Chen, Xia, Zhao, Zhang, Yu, Liang, and
  Zhu}]{lu2021iconqa}
Lu, P.; Qiu, L.; Chen, J.; Xia, T.; Zhao, Y.; Zhang, W.; Yu, Z.; Liang, X.; and
  Zhu, S.-C. 2021.
\newblock Iconqa: A new benchmark for abstract diagram understanding and visual
  language reasoning.
\newblock \emph{arXiv preprint arXiv:2110.13214}.

\bibitem[{Peng, Zeng, and Zhao(2022)}]{DBLP:conf/emnlp/PengZZ22}
Peng, R.; Zeng, Y.; and Zhao, J. 2022.
\newblock Distill The Image to Nowhere: Inversion Knowledge Distillation for
  Multimodal Machine Translation.
\newblock In Goldberg, Y.; Kozareva, Z.; and Zhang, Y., eds., \emph{Proceedings
  of the 2022 Conference on Empirical Methods in Natural Language Processing,
  {EMNLP} 2022, Abu Dhabi, United Arab Emirates, December 7-11, 2022},
  2379--2390. Association for Computational Linguistics.

\bibitem[{Plummer et~al.(2017)Plummer, Wang, Cervantes, Caicedo, Hockenmaier,
  and Lazebnik}]{DBLP:journals/ijcv/PlummerWCCHL17}
Plummer, B.~A.; Wang, L.; Cervantes, C.~M.; Caicedo, J.~C.; Hockenmaier, J.;
  and Lazebnik, S. 2017.
\newblock Flickr30k Entities: Collecting Region-to-Phrase Correspondences for
  Richer Image-to-Sentence Models.
\newblock \emph{Int. J. Comput. Vis.}, 123(1): 74--93.

\bibitem[{Radford et~al.(2021)Radford, Kim, Hallacy, Ramesh, Goh, Agarwal,
  Sastry, Askell, Mishkin, Clark, Krueger, and
  Sutskever}]{DBLP:conf/icml/RadfordKHRGASAM21}
Radford, A.; Kim, J.~W.; Hallacy, C.; Ramesh, A.; Goh, G.; Agarwal, S.; Sastry,
  G.; Askell, A.; Mishkin, P.; Clark, J.; Krueger, G.; and Sutskever, I. 2021.
\newblock Learning Transferable Visual Models From Natural Language
  Supervision.
\newblock In Meila, M.; and Zhang, T., eds., \emph{Proceedings of the 38th
  International Conference on Machine Learning, {ICML} 2021, 18-24 July 2021,
  Virtual Event}, volume 139 of \emph{Proceedings of Machine Learning
  Research}, 8748--8763. {PMLR}.

\bibitem[{Raffel et~al.(2020)Raffel, Shazeer, Roberts, Lee, Narang, Matena,
  Zhou, Li, and Liu}]{DBLP:journals/jmlr/RaffelSRLNMZLL20}
Raffel, C.; Shazeer, N.; Roberts, A.; Lee, K.; Narang, S.; Matena, M.; Zhou,
  Y.; Li, W.; and Liu, P.~J. 2020.
\newblock Exploring the Limits of Transfer Learning with a Unified Text-to-Text
  Transformer.
\newblock \emph{J. Mach. Learn. Res.}, 21: 140:1--140:67.

\bibitem[{Rombach et~al.(2022)Rombach, Blattmann, Lorenz, Esser, and
  Ommer}]{DBLP:conf/cvpr/RombachBLEO22}
Rombach, R.; Blattmann, A.; Lorenz, D.; Esser, P.; and Ommer, B. 2022.
\newblock High-Resolution Image Synthesis with Latent Diffusion Models.
\newblock In \emph{{IEEE/CVF} Conference on Computer Vision and Pattern
  Recognition, {CVPR} 2022, New Orleans, LA, USA, June 18-24, 2022},
  10674--10685. {IEEE}.

\bibitem[{Ronneberger, Fischer, and
  Brox(2015)}]{DBLP:conf/miccai/RonnebergerFB15}
Ronneberger, O.; Fischer, P.; and Brox, T. 2015.
\newblock U-Net: Convolutional Networks for Biomedical Image Segmentation.
\newblock In Navab, N.; Hornegger, J.; III, W. M.~W.; and Frangi, A.~F., eds.,
  \emph{Medical Image Computing and Computer-Assisted Intervention - {MICCAI}
  2015 - 18th International Conference Munich, Germany, October 5 - 9, 2015,
  Proceedings, Part {III}}, volume 9351 of \emph{Lecture Notes in Computer
  Science}, 234--241. Springer.

\bibitem[{Song, Meng, and Ermon(2021)}]{DBLP:conf/iclr/SongME21}
Song, J.; Meng, C.; and Ermon, S. 2021.
\newblock Denoising Diffusion Implicit Models.
\newblock In \emph{9th International Conference on Learning Representations,
  {ICLR} 2021, Virtual Event, Austria, May 3-7, 2021}. OpenReview.net.

\bibitem[{Vaswani et~al.(2017)Vaswani, Shazeer, Parmar, Uszkoreit, Jones,
  Gomez, Kaiser, and Polosukhin}]{vaswani2017attention}
Vaswani, A.; Shazeer, N.; Parmar, N.; Uszkoreit, J.; Jones, L.; Gomez, A.~N.;
  Kaiser, {\L}.; and Polosukhin, I. 2017.
\newblock Attention is all you need.
\newblock In \emph{Advances in neural information processing systems},
  5998--6008.

\bibitem[{Wei et~al.(2022)Wei, Wang, Schuurmans, Bosma, Chi, Le, and
  Zhou}]{DBLP:journals/corr/abs-2201-11903}
Wei, J.; Wang, X.; Schuurmans, D.; Bosma, M.; Chi, E.~H.; Le, Q.; and Zhou, D.
  2022.
\newblock Chain of Thought Prompting Elicits Reasoning in Large Language
  Models.
\newblock \emph{CoRR}, abs/2201.11903.

\bibitem[{Wu et~al.(2021)Wu, Kong, Bi, Li, and Kao}]{wu2021good}
Wu, Z.; Kong, L.; Bi, W.; Li, X.; and Kao, B. 2021.
\newblock Good for Misconceived Reasons: An Empirical Revisiting on the Need
  for Visual Context in Multimodal Machine Translation.
\newblock \emph{arXiv preprint arXiv:2105.14462}.

\bibitem[{Xue et~al.(2021)Xue, Constant, Roberts, Kale, Al{-}Rfou, Siddhant,
  Barua, and Raffel}]{DBLP:conf/naacl/XueCRKASBR21}
Xue, L.; Constant, N.; Roberts, A.; Kale, M.; Al{-}Rfou, R.; Siddhant, A.;
  Barua, A.; and Raffel, C. 2021.
\newblock mT5: {A} Massively Multilingual Pre-trained Text-to-Text Transformer.
\newblock In Toutanova, K.; Rumshisky, A.; Zettlemoyer, L.;
  Hakkani{-}T{\"{u}}r, D.; Beltagy, I.; Bethard, S.; Cotterell, R.;
  Chakraborty, T.; and Zhou, Y., eds., \emph{Proceedings of the 2021 Conference
  of the North American Chapter of the Association for Computational
  Linguistics: Human Language Technologies, {NAACL-HLT} 2021, Online, June
  6-11, 2021}, 483--498. Association for Computational Linguistics.

\bibitem[{Yin et~al.(2020)Yin, Meng, Su, Zhou, Yang, Zhou, and
  Luo}]{yin-etal-2020-novel}
Yin, Y.; Meng, F.; Su, J.; Zhou, C.; Yang, Z.; Zhou, J.; and Luo, J. 2020.
\newblock A Novel Graph-based Multi-modal Fusion Encoder for Neural Machine
  Translation.
\newblock In \emph{Proceedings of the 58th Annual Meeting of the Association
  for Computational Linguistics}, 3025--3035.

\bibitem[{Yu et~al.(2019)Yu, Yu, Cui, Tao, and Tian}]{yu2019deep}
Yu, Z.; Yu, J.; Cui, Y.; Tao, D.; and Tian, Q. 2019.
\newblock Deep modular co-attention networks for visual question answering.
\newblock In \emph{Proceedings of the IEEE/CVF conference on computer vision
  and pattern recognition}, 6281--6290.

\bibitem[{Zhang et~al.(2020)Zhang, Chen, Wang, Utiyama, Sumita, Li, and
  Zhao}]{Zhang2020Neural}
Zhang, Z.; Chen, K.; Wang, R.; Utiyama, M.; Sumita, E.; Li, Z.; and Zhao, H.
  2020.
\newblock Neural Machine Translation with Universal Visual Representation.
\newblock In \emph{International Conference on Learning Representations}.

\bibitem[{Zhang et~al.(2023{\natexlab{a}})Zhang, Chen, Wang, Utiyama, Sumita,
  Li, and Zhao}]{DBLP:journals/corr/abs-2301-03344}
Zhang, Z.; Chen, K.; Wang, R.; Utiyama, M.; Sumita, E.; Li, Z.; and Zhao, H.
  2023{\natexlab{a}}.
\newblock Universal Multimodal Representation for Language Understanding.
\newblock \emph{CoRR}, abs/2301.03344.

\bibitem[{Zhang et~al.(2022)Zhang, Zhang, Li, and
  Smola}]{DBLP:journals/corr/abs-2210-03493}
Zhang, Z.; Zhang, A.; Li, M.; and Smola, A. 2022.
\newblock Automatic Chain of Thought Prompting in Large Language Models.
\newblock \emph{CoRR}, abs/2210.03493.

\bibitem[{Zhang et~al.(2023{\natexlab{b}})Zhang, Zhang, Li, Zhao, Karypis, and
  Smola}]{DBLP:journals/corr/abs-2302-00923}
Zhang, Z.; Zhang, A.; Li, M.; Zhao, H.; Karypis, G.; and Smola, A.
  2023{\natexlab{b}}.
\newblock Multimodal Chain-of-Thought Reasoning in Language Models.
\newblock \emph{CoRR}, abs/2302.00923.

\end{thebibliography}

\clearpage

\appendix

\section{Appendix}

\begin{algorithm}
    \caption{Multi-modal Latent Space Learning for CoT}
    \begin{algorithmic}[1] %每行显示行号
    \REQUIRE Language Context $X_{L}$ Question $X_Q$, Options $X_O$, Image Input $X_{V}$
    \ENSURE Rationale $\hat{Y_R}$, Answer $\hat{Y_A}$
    \STATE Text Representation: \\
    $Z_L^R \gets \textsc{Encoder}_{text}([X_L; X_Q; X_O])$
    \STATE Multi-modal Latent Vector: \\
    $Z_{V}^R \gets \textsc{Diffusion}(X_{V}, Z_L^R)$
    \STATE Feature Fusion: \\
    $Z_{fuse}^R \gets \textsc{Fuse}(Z_{V}^R, Z_L^R)$
    \STATE Rationale Generation: \\
    $\hat{Y_R} \gets \textsc{Decoder}_{text}(Z_{fuse}^R)$
    \STATE Text Representation: \\
    $Z_L^A \gets \textsc{Encoder}_{text}([X_L; X_Q; X_O; \hat{Y_R}])$
    \STATE Multi-modal Latent Vector: \\
    $Z_{V}^A \gets \textsc{Diffusion}(X_{V}, Z_L^A)$
    \STATE Feature Fusion: \\
    $Z_{fuse}^A \gets \textsc{Fuse}(Z_{V}^A, Z_L^A)$
    \STATE Answer Inference: \\
    $\hat{Y_A} \gets \textsc{Decoder}_{text}(Z_{fuse}^A)$
    \end{algorithmic}
    \label{alg:method}
\end{algorithm}

\subsection{A. Chain-of-Thought}
\label{sec:CoT}

There are two primary paradigms that large language models (LLMs) use to perform CoT reasoning: Zero-shot-CoT~\cite{DBLP:journals/corr/abs-2205-11916} and Few-shot-CoT~\cite{DBLP:journals/corr/abs-2201-11903}. Zero-Shot-CoT demonstrates that LLMs can reason effectively without any prior training by using a simple prompt such as "Let's think step by step" to facilitate systematic thinking before answering a question. On the other hand, Few-Shot-CoT uses a few demonstrations, each consisting of a question and a reasoning chain leading to an answer. Depending on the method of obtaining these demonstrations, whether through manual crafting or automatic generation, they are classified as Manual-CoT~\cite{DBLP:journals/corr/abs-2201-11903} or Auto-CoT~\cite{DBLP:journals/corr/abs-2210-03493}. Few-shot-CoT employs reasoning demonstrations one-by-one. For instance, fine-tuning PaLM 540B with eight CoT exemplars achieves state-of-the-art accuracy on the GSM8K benchmark for math word problems, surpassing even finetuned GPT-3 with a verifier \cite{DBLP:journals/corr/abs-2201-11903}.

Fine-tuning language models is another approach that has demonstrated significant performance in eliciting Chain-of-Thought (CoT) reasoning. Lu et al.~\cite{DBLP:journals/corr/abs-2209-09513} introduced Science Question Answering (ScienceQA), a new benchmark consisting of approximately 21,000 multi-modal multiple-choice questions with annotations of their answers and corresponding lectures and explanations. They fine-tuned language models on this dataset and achieved a 1.20\% improvement in few-shot GPT-3's question-answering performance. Moreover, prior studies indicate that CoT reasoning abilities may emerge in LLMs with over 100 billion parameters. In other words, language models with fewer than 100 billion parameters tend to generate illogical rationales, also known as hallucinations.

\subsection{B. MM-CoT}
\label{sec:mmCoT}

MM-CoT uses text representation encoded by a large language model and fixed image features encoded by off-the-shelf vision extraction models, fused via an attention mechanism. In its implementation, it adopts a two-stage framework consisting of rationale generation and answer inference. Experimental results demonstrate that this framework performs better on question answering tasks than generating rationales and answers together.

Despite recent advances, there are still challenges in using the visual modality more effectively. Firstly, most off-the-shelf vision extraction models are not specifically trained for complex CoT reasoning. Secondly, when a two-stage framework is used, fixed image features may not be suitable for different stages and flexible language queries. It is our belief that fixed and shallow vision features may not be the optimal choice to enhance language models' complex reasoning abilities and performance on question answering tasks. As shown in Figure \ref{fig:example}, the shallow vision features just include information about grass, football, slogan sign and so on. But we need language models to generate more complex reasoning such as "This child should play football on the playground". So our work aims at learning flexible and deep vision features used to improve complex reasoning ability.

\subsection{C. Diffusion Models}

The diffusion model is a method inspired by the physical process of gas diffusion, which has found applications in various scientific fields. In recent years, it has shown impressive results in image generation and become the new state-of-the-art in deep generative models, surpassing the original leader, GAN~\cite{DBLP:conf/nips/GoodfellowPMXWOCB14}, for image generation tasks.
Typically, diffusion models involve two processes: the forward process (also known as the diffusion process) and the reverse process. Specifically, Denoising Diffusion Probabilistic Models (DDPMs)~\cite{DBLP:conf/nips/HoJA20} disrupt images by adding noise through multiple steps of the forward process and subsequently generate samples by gradually removing the noise through multiple reverse steps. However, DDPMs process image information at the pixel level, and require simulating a Markov chain for many steps to produce a sample, resulting in slow sampling speed.

Subsequently, most research on diffusion models has focused on accelerating the speed of sample generation. Denoising Diffusion Implicit Models (DDIMs)~\cite{DBLP:conf/iclr/SongME21} are more efficient iterative implicit probabilistic models with the same training objective as DDPMs, but use non-Markovian diffusion processes. Stable diffusion~\cite{DBLP:conf/cvpr/RombachBLEO22} runs the diffusion process on the latent space to accelerate the processing of image information. The diffusion process occurs within the "image information creator" component. Given token embeddings that represent input text and a randomly generated starting image information array (also known as latents), the process produces an information array that the image decoder uses to generate the final image. This excellent generative performance comes from the understanding of the flexible and deep vision features aligned with text representation.
Existing studies have demonstrated that diffusion models already have a semantic latent space~\cite{DBLP:journals/corr/abs-2210-10960}. Inspired by this idea, we further propose a multi-modal latent space learning approach via the diffusion process for chain-of-thought reasoning in language models.

\begin{figure}[h] %插入图片
\centering %图片居中
\pgfplotsset{height=5.2cm,width=7.68cm,compat=1.14,every axis/.append style={thick},every tick label/.append style={font=\small},every axis legend/.append style={ at={(0.5,0.95)}},legend columns=2 row=1} 

\begin{tikzpicture} %tikz图片
%\tikzset{every node}=[font=\small]
\begin{axis}[
                align = center,
                width = 7.68cm,
                height=5.2cm, 
                legend style={at={(0.5,-0.2)},anchor=north},
                grid=major,
                ymin=80, ymax=95,
                xticklabels={0.15,0.16,0.17,0.18,0.19,0.2}, 
                xtick={0.15,0.16,0.17,0.18,0.19,0.2},
                ylabel style={align=center},
                xlabel={Scale-factor of Latent},
                ylabel={Accuracy},
                ytick={80,85,90,95},
                ymajorgrids=true,
                xmajorgrids=true,
                grid style=dashed,
                xtick pos=bottom,
                ytick pos=left,
                %legend pos=north west
                %tick align=outside, %刻度在外显式
                legend pos=north west
    ]

%第一条线，mark是折线标示形状
\addplot[mark=o,red] plot coordinates { 
    (0.15215, 90.40)
    (0.16215, 90.59)
    (0.17215, 90.71)
    (0.18215, 90.97)
    (0.19215, 90.36)
    (0.20215, 89.93)
};

%图例里名字
\addlegendentry{DPMM-CoT}

\addplot[mark=triangle,blue] plot coordinates { 
    (0.15215, 84.91)
    (0.16215, 84.91)
    (0.17215, 84.91)
    (0.18215, 84.91)
    (0.19215, 84.91)
    (0.20215, 84.91)
};

%图例里名字
\addlegendentry{MM-CoT}

\end{axis}
\end{tikzpicture}
\caption{Accuracy curve of DPMM-CoT across scale-factor of latent.}
\label{fig:latent}
\end{figure}
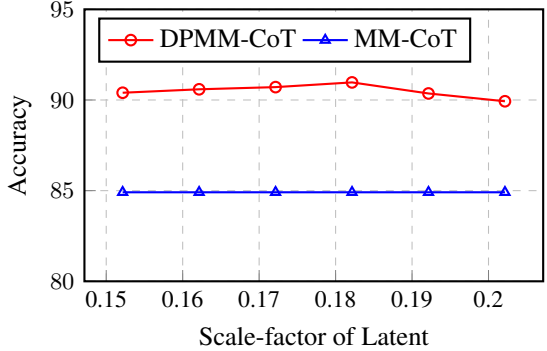

\subsection{D. Introduction to Datasets}
\subsubsection{D.1 ScienceQA}

This dataset comprises approximately 21,000 multi-modal multiple-choice questions from various scientific fields. Each example includes a question, potential answers, a correct answer, context, and a series of explanations and solutions. The dataset covers three subjects, 26 topics, 127 categories, and 379 skills. It is divided into training, validation, and test sets, containing 12,726, 4,241, and 4,241 examples, respectively. Note that some images in the dataset come with captions; however, we chose not to use them in our model since their inclusion for image replacement can result in significant information loss.

\subsubsection{D.2 Multi-modal Translation Dataset}

This dataset extends the
Flickr30K dataset~\cite{DBLP:journals/ijcv/PlummerWCCHL17} which consists of 30K images and 150K descriptive captions. And Multi30K comprises translations created by professional translators over a subset of the English descriptions. We conducted experiments on English-to-German and English-to-French translations, using three test sets: test2016-flickr (Test16), test2017-flickr (Test17), and test2017-mscoco (MSCOCO) containing 1000, 1000, 461 examples, respectively.

\subsection{E. Settings}

In our experiment on CoT in LLMs, we employed a two-stage framework consisting of two procedures: rationale generation and answer inference. Both stages shared the same model architecture, namely the T5 encoder-decoder architecture~\cite{DBLP:journals/jmlr/RaffelSRLNMZLL20}. The model was initialized using the pre-trained UnifiedQA~\cite{khashabi-etal-2020-unifiedqa} checkpoint in the same manner for both stages. Additionally, we initialized the VAE and UNet modules in the multi-modal latent space learning with checkpoints from Stable-Diffusion-v1-4~\cite{DBLP:conf/cvpr/RombachBLEO22}. The model was then fine-tuned for 20 epochs with joint sequence generation loss and latent diffusion model loss, where the maximum length of the input sequence was set to 512. The length of the output sequence was set to 512 and 64 for the stage of rationale generation and answer inference respectively. We used the AdamW optimizer with an initial learning rate of 5e-5 to optimize our model. Furthermore, we used ROUGE-L~\cite{lin2003rouge} as a metric to evaluate the quality of generated rationales and accuracy to evaluate inferred answers.
For our experiment on multi-modal machine translation (MMT), we employed the mT5 encoder-decoder architecture, which was initialized using the pre-trained mT5-large~\cite{DBLP:conf/naacl/XueCRKASBR21} checkpoint, which had been pre-trained on a multilingual corpus. We also initialized the VAE and UNet modules in a similar way to the CoT experiment in LLMs. We fine-tuned the MMT model for 10 epochs, using an initial learning rate of 5e-5. To evaluate translation performance, we used tokenized BLEU scores with the NLTK~\cite{Bird_Natural_Language_Processing_2009} tokenizer.

\subsection{F. Influence of Scale Factor $\eta$}
\label{sec:scale factor}
The scale factor was introduced in the High-Resolution Image Synthesis With Latent~\cite{DBLP:conf/cvpr/RombachBLEO22}. The goal was to handle different latent spaces (from different autoencoders, which can be scaled quite differently than images) with similar noise schedules. The scale factor ensures that the initial latent space on which the diffusion model is operating has approximately unit variance. 

In DPMM-CoT, we use VAE as the autoencoder, and therefore we introduce the scale factor $\tau$ to rescale the produced latents before feeding them to the UNet. In a previous study~\cite{DBLP:conf/cvpr/RombachBLEO22}, researchers determined the optimal scale factor for ensuring unit variance across all dimensions by averaging over examples generated by the VAE. They found that the suitable value of $\tau$ was 0.18215.

To evaluate the rationality of this value in our DPMM-CoT, we conducted experiments to compare the effects of different scale factors on our results. We tested values of $\tau$ ranging between 0.15215 and 0.20215.

The experimental results, shown in Figure \ref{fig:latent}, yielded accuracy rates of 90.40\%, 90.59\%, 90.71\%, 90.97\%, 90.36\%, and 89.93\% for the respective values of $\tau$. We found that DPMM-CoT$_\textrm{Base}$ achieves the highest accuracy when $\tau$ was set to 0.18215.

Furthermore, the accuracy decreased with greater absolute differences between the scale factor $\tau$ and 0.18215. This trend is evident from the accuracy curve, which suggests that 0.18215 is the ideal scale factor value for ensuring unit variance in the VAE latents.

\begin{figure}[t]
\centering
\includegraphics[width=0.45\textwidth]
{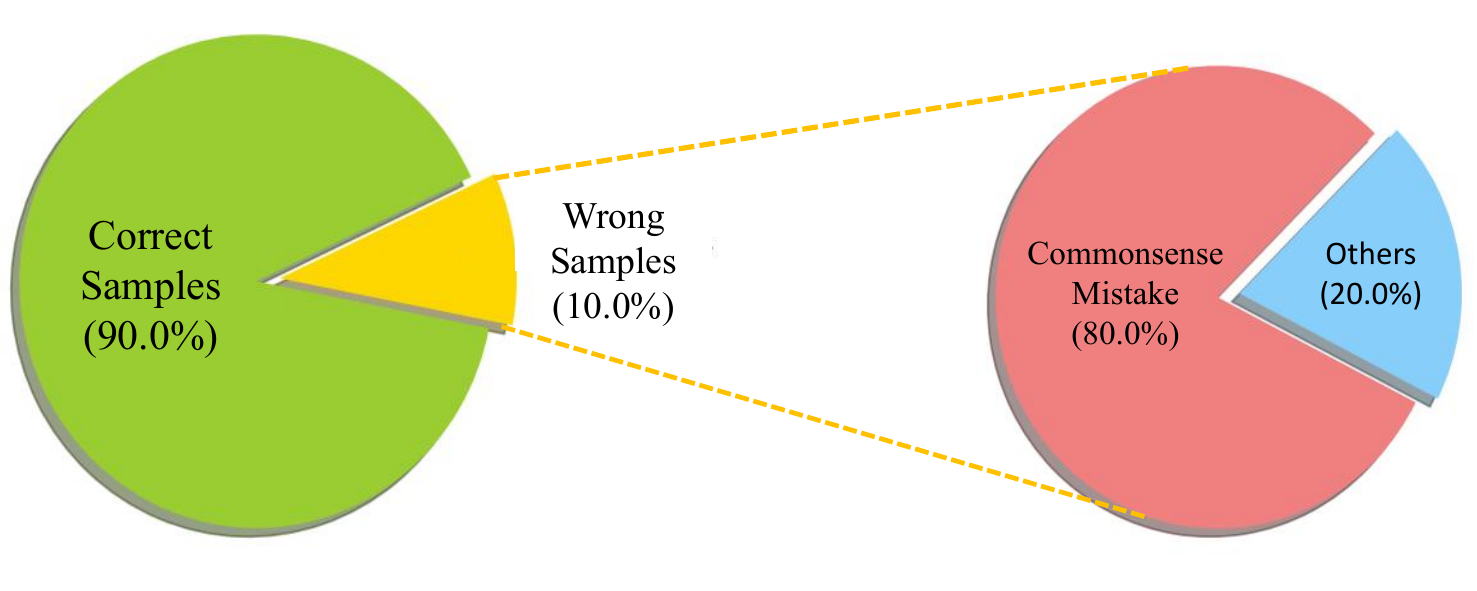}
\caption{Statistical study of random samples.}
\label{fig:statistic}
\end{figure}

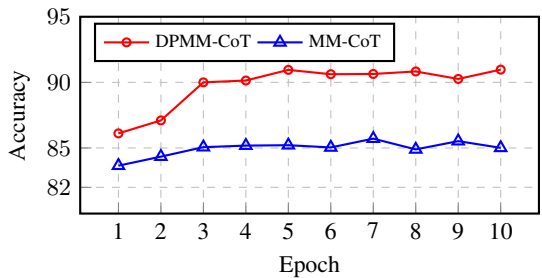
\begin{figure}[t]
	\centering
	\pgfplotsset{height=5.2cm,width=7.68cm,compat=1.14,every axis/.append style={thick},every tick label/.append style={font=\small},every axis legend/.append style={ at={(0.5,0.95)}},legend columns=2 row=1} 
\begin{tikzpicture} \tikzset{every node}=[font=\small] 
\begin{axis} [
                align = center,
                width = 7.68cm,
                height=4.2cm, 
                legend style={at={(0.5,1.3)},anchor=north},
                ymin=80, ymax=95,
                xticklabels={1,2,3,4,5,6,7,8,9,10}, xtick={0,1,2,3,4,5,6,7,8,9},
                ylabel style={align=center},
                xlabel={Epoch},
                ylabel={Accuracy},
                ytick={82, 85, 90, 95},
                ymajorgrids=true,
                xmajorgrids=true,
                grid style=dashed,
                xtick pos=bottom,
                ytick pos=left,
                legend pos=north west
                ]

\addplot+[color=red,
                    mark=o,
                    mark size=1.5pt,
                    ] coordinates {(0, 86.11) (1, 87.10) (2, 90.00) (3, 90.14) (4, 90.95) (5, 90.62) (6, 90.64) (7, 90.83) (8, 90.26) (9, 90.97)};%\label{plot_2}
\addlegendentry{\scriptsize DPMM-CoT}

\addplot+ [color=blue,
                    mark=triangle,
                    mark size=2.5pt,
                    ] coordinates { (0, 83.65)  (1, 84.34) (2, 85.06) (3, 85.18) (4, 85.21) (5, 85.04) (6, 85.71) (7, 84.89) (8, 85.52) (9, 85.01)};%\label{plot_1}
\addlegendentry{\scriptsize MM-CoT}

\end{axis}
\end{tikzpicture}
\vspace{-3mm}
    \caption{Accuracy curve of the MM-CoT and DPMM-CoT across epochs.}
    \label{fig:epoch}
 \vspace{-5mm}
\end{figure}

\subsection{G. Influence of Quality of Rationale}
\label{sec:quality of rationale}
Table \ref{tab:rationale} shows a comparison of MM-CoT$_\textrm{Base}$ and DPMM-CoT$_\textrm{Base}$ in both the two-stage framework. ROUGE-L measures the performance in rationale generation, while accuracy measures the performance in answer inference. The results show that DPMM-CoT$_\textrm{Base}$ surpasses MM-CoT Base by 1.21 in rationale generation and by 6.06\% in answer inference. These observations illustrates the effectiveness of multi-modal latent space learning via diffusion process.
It's important to note that the ROUGE-L score may not fully reflect the correctness of the rationale. Figure \ref{fig:case_analyse} illustrates an example where the rationale generated by both MM-CoT and DPMM-CoT are similar, but the difference in the last sentence between "logos" and "pathos" leads to completely different answers. Hence, high scores of ROUGE-L may be insufficient to demonstrate the complex reasoning ability of language models as those models may only be simulating standard rationale without understanding the language thought of text context and image context of problems. 

\subsection{H. Stability of Improvements}

To better demonstrate the confidence level of our method's performance compared to MM-CoT, we visualized the evaluation accuracy curve of both models in different training epochs under the same conditions. Figure \ref{fig:epoch} showcases our proposed multi-modal latent space learning's ability to bring stable benefits throughout the entire training process, rather than just specific moments, as evidenced by the improved performance in all epochs compared to the baseline.

\begin{table}[t]
    \small
    \setlength\tabcolsep{2pt}
    \centering
    \caption{BLEU score of EN-DE and EN-FR tasks. Encouragingly, our DPMM-MT is an image-must MMT model.\\
    Transformer\cite{vaswani2017attention};
    Fusion-conv\cite{caglayan-etal-2017-lium};
    Trg-mul\cite{caglayan-etal-2017-lium};
    UVR-NMT\cite{Zhang2020Neural};
    GMNMT\cite{yin-etal-2020-novel};
    DCCN\cite{lin2020dynamic};
    ImagiT \cite{long-etal-2021-generative};
    Gated Fusion\cite{wu2021good};
    RMMT\cite{wu2021good};
    IKD-MMT \cite{DBLP:conf/emnlp/PengZZ22}
    }
    \label{tab:translation_detail}
    \vspace{-3mm}
    {\begin{tabular}{lccccc}
    \toprule
    \multicolumn{1}{c}{\multirow{2}*{\textbf{Model}}} & \multicolumn{3}{c}{\textbf{EN-DE}} & \multicolumn{2}{c}{\textbf{EN-FR}} \\
    \cmidrule(lr){2-4} \cmidrule(lr){5-6}
    & \multicolumn{1}{c}{\textbf{Test16}} & \multicolumn{1}{c}{\textbf{Test17}} & \multicolumn{1}{c}{\textbf{MSCOCO}} & \multicolumn{1}{c}{\textbf{Test16}} & \multicolumn{1}{c}{\textbf{Test17}} \\
    \midrule
    % \multicolumn{1}{c}{\textit{Language-only MMT}}\\
    Transformer & 37.6 & 31.7 & 27.9 & 59.0 & 51.9  \\
    % Multitask\cite{elliott-kadar-2017-imagination} & 36.8 & - & - & - & -  \\
    % NMT$_{\text{SRC+IMG}}$\cite{DBLP:conf/acl/CalixtoLC17} & 36.5 & - & - & - & - \\
    % IMG$_D$\cite{DBLP:conf/emnlp/CalixtoL17} & 37.3 & - & - & - & -\\
    Fusion-conv & 37.0 & 29.8 & 25.1 & 53.5 & 51.6\\
    Trg-mul & 37.8 & 30.7 & 26.4 & 54.7 & 52.7\\
    % VAG-NMT\cite{zhou-etal-2018-visual} & - & 31.6 & 28.3 & - & 53.8 \\
    % VMMT$_\text{F}$\cite{calixto-etal-2019-latent} & 37.7 & 30.1 & 25.5 & - & - \\
    % DS-SUM-L2\cite{caglayan2019multimodal} & 39.4 & 32.6 & - & 60.7 & 54.2 \\
    % Del+obj\cite{ive-etal-2019-distilling} & 38.0 & - & - & 59.8 & - \\
    UVR-NMT & 36.94 & 28.63 & - & 57.53 & 48.46 \\
    % Multimodal\cite{yao-wan-2020-multimodal} & 38.7 & - & - & - & - \\
    GMNMT & 39.8 & 32.2 & 28.7 & 60.9 & 53.9 \\
    DCCN & 39.7 & 31.0 & 26.7 & 61.2 & 54.3 \\
    ImagiT & 38.5 & 32.1 & 28.7 & 59.7 & 52.4 \\

    % \midrule
    % \multicolumn{1}{c}{\textit{Multi-modal MMT}}\\
    
    % Gumbel-att\cite{liu2021gumbel} & 39.2 & 31.4 & 26.9 & - & - \\
    % OVC+${L}_{m}$\cite{wang2021efficient} & - & 32.3 & 28.9 & - & 54.1 \\
    Gated Fusion & \textbf{41.96} & 33.59 & 29.04 & 61.69 & 54.85 \\
    RMMT & 41.45 & 32.94 & 30.01 & 62.1 & 54.39 \\
    IKD-MMT  & 41.28 & 33.83 & 30.17 & 62.53 & 54.84 \\
    \hdashline
    {mT5} & 38.56 & {33.01} & {28.10} & {61.71} & {53.84} \\
    \textbf{DPMM-MT} & 41.63 & \textbf{36.18} & \textbf{30.75} & \textbf{66.91} & \textbf{57.80} \\
    \bottomrule
\end{tabular}}
\end{table}

\subsection{I. Case Study}
\label{sec:case study}
To perform an in-depth analysis of the effectiveness of DPMM-CoT, our study involved randomly selecting and analyzing 100 test set samples one by one. Out of these, we found that 90 were correct, while 10 were incorrect. Notably, 80\% of the incorrect samples required background knowledge on geography and alphabet, which was not provided in the text or image context. These statistics are presented in Figure \ref{fig:statistic}. Our findings indicate that DPMM-CoT has great performance on ScienceQA tasks, owing to its ability to effectively leverage multi-modal information for generating useful rationale and inferring correct answers. 

\begin{figure*}[t]
  \centering
  \includegraphics[width=1\textwidth]{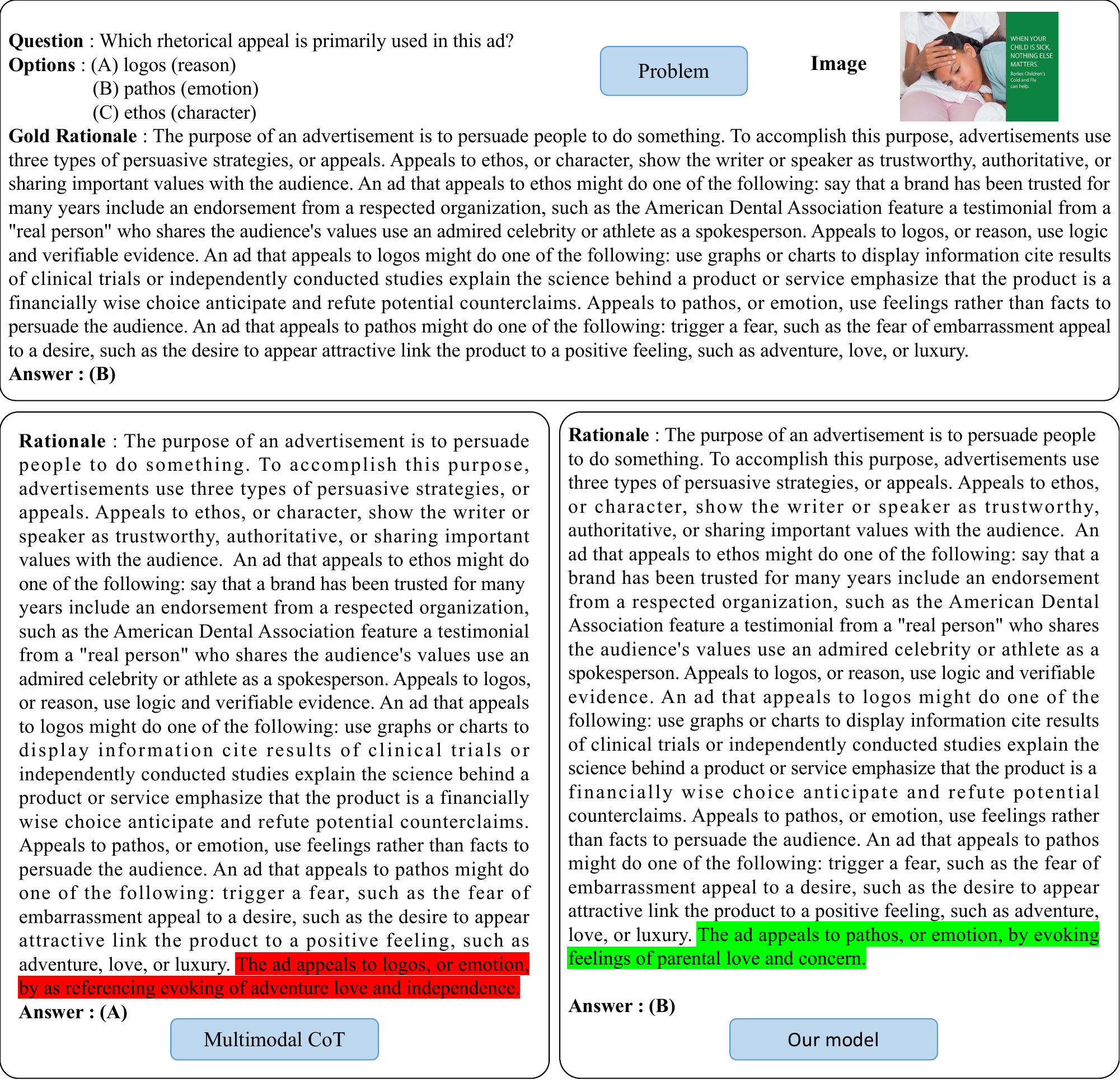}
  \caption{This is a question about how to determine how advertisements impress consumers. The result shows that our model can understand the deep-level information of image. It can infer the the emotions hidden behind the image about a sick girl. And it can finally choose the right answer.}\label{fig:case_analyse}
  
\end{figure*}

We conducted an analysis of a specific case to demonstrate the effectiveness of multi-modal latent space learning via diffusion process. The results, as shown in Figure \ref{fig:case_analyse}, reveal that DPMM-CoT is capable of generating useful rationale and inferring the correct answer due to its deep-level comprehension of the flexible and aligned image features, whereas MM-CoT cannot. The case in question concerns determining the most effective means by which advertisements can impress consumers, with three options provided: (A) logos (reason), (B) pathos (emotion), and (C) ethos (character). Logos employs verifiable evidence and logical reasoning to persuade individuals to take action, while pathos relies on emotions rather than facts. Ethos demonstrates the speaker or writer as trustworthy, authoritative, or sharing key values with their audience. To infer the correct answer in this scenario, it is insufficient to simply understand the text's context and identify the presence of a girl in the image. The model must also comprehend that the girl is feeling exhausted and that the caption in the image highlights the importance of parental love and concern for sick children. The rationale produced by DPMM-CoT indicates that it possesses a deep understanding of language thought of text and image context, whereas the correct answer illustrates that our model can perceive the emotions conveyed by the image of a sick girl. This case serves as a compelling demonstration of DPMM-CoT's exceptional ability to combine multi-modal information and enhance a language model's intricate reasoning capabilities, all thanks to the efficacy of multi-modal latent space learning via diffusion process.

% [1页]

% \begin{table}[ht]
% \resizebox{\linewidth}{!}{
%     \centering\small
%     \begin{tabular}{lrrrrr} \toprule
%     & Train & Dev & Test2016 &  Test2017 & MSCOCO &\\ 
%       \midrule

%     Multi30K &  29K &  1014 &  1000 & 1000 & 461 \\
%      \bottomrule
    
%     \end{tabular}
%     }
%     \caption{Parallel corpus sizes.}
%     \label{tab:opus_data}
% \end{table}

\end{document}